\documentclass[pmlr,twocolumn,10pt]{jmlr} %

\usepackage{booktabs}
\usepackage{siunitx}

\usepackage[switch]{lineno}
\usepackage{xcolor}
\usepackage{soul}
\usepackage{dblfloatfix}
\usepackage{placeins}

\soulregister\ref7
\soulregister\eqref7
\soulregister\cite7
\soulregister\citep7
\soulregister\citet7
\soulregister\citealt7
\soulregister\citeauthor7
\soulregister\section7
\soulregister\subsection7
\soulregister\paragraph7

\definecolor{lightyellow}{cmyk}{0,0,0.5,0}
\sethlcolor{lightyellow}

\theorembodyfont{\upshape}
\theoremheaderfont{\scshape}
\theorempostheader{:}
\theoremsep{\newline}

\jmlrvolume{287}
\jmlryear{2025}
\jmlrsubmitted{} %
\jmlrpublished{} %
\jmlrworkshop{Conference on Health, Inference, and Learning (CHIL) 2025}

\title[Diagnosing our datasets: How does my language model learn clinical information?]{Diagnosing our datasets:\\ How does my language model learn clinical information?}

\author{%
\Name{Furong Jia} \Email{flora.jia@duke.edu}\\
\addr Duke University, United States
\AND
\Name{David Sontag} \Email{dsontag@csail.mit.edu}\\
\addr MIT CSAIL, United States
\AND
\Name{Monica Agrawal} \Email{monica.agrawal@duke.edu}\\
\addr Duke University, United States
}

\newcommand{\datasetname}{\texttt{MedLingo}}
\usepackage{multirow}  %

\begin{document}

\maketitle

\begin{abstract}
Large language models (LLMs) have performed well across various clinical natural language processing tasks, despite not being directly trained on electronic health record (EHR) data. In this work, we examine how popular open-source LLMs learn clinical information from large mined corpora through two crucial but understudied lenses: (1) their interpretation of clinical jargon, a foundational ability for understanding real-world clinical notes, and (2) their responses to unsupported medical claims. For both use cases, we investigate the frequency of relevant clinical information in their corresponding pretraining corpora, the relationship between pretraining data composition and model outputs, and the sources underlying this data. To isolate clinical jargon understanding, we evaluate LLMs on a new dataset \datasetname{}. Unsurprisingly, we find that the frequency of clinical jargon mentions across major pretraining corpora correlates with model performance. However, jargon frequently appearing in clinical notes often rarely appears in pretraining corpora, revealing a mismatch between available data and real-world usage. Similarly, we find that a non-negligible portion of documents support disputed claims that can then be parroted by models. Finally, we classified and analyzed the types of online sources in which clinical jargon and unsupported medical claims appear, with implications for future dataset composition. 
\end{abstract}

\paragraph*{Data and Code Availability}

This paper leverages publicly available pre-training corpora, the Clinical Acronym Sense Inventory (CASI) dataset, and the MIMIC-IV dataset \citep{moon2014sense, johnson2023mimic, johnson2020mimic}. 
The code, our new benchmark \datasetname{}, and analysis results can be found here: 
\url{https://github.com/Flora-jia-jfr/diagnosing_our_datasets}

\paragraph*{Institutional Review Board (IRB)}
This research does not require IRB approval.

\section{Introduction}

In recent years, there has been significant warranted excitement around the application of large language models (LLMs) to diverse clinical applications, including information extraction, summarization, question answering, and trial matching \citep{li2024scoping, van2024adapted, agrawal-etal-2022-large, zakka2024almanac, jin2024matching}. Researchers have found promising performance with both off-the-shelf general domain models (e.g., GPT and Llama families), as well as models fine-tuned specifically with biomedical corpora, such as PubMed, clinical guidelines, and medical question answering datasets ~\citep{chen2023meditron, christophe2024med42}. 
Recent research has actually found that general domain models can perform just as well as these medically fine-tuned counterparts on standard benchmarks, despite being trained only on general online corpora \citep{jeong2024medical, li2024clinicalt5modelsbetter}. This raises major questions around where and how open-source LLMs are learning clinical information, given that they are not trained on EHR text.

\begin{figure*}[htp]
    \centering
    \includegraphics[width=1\textwidth]{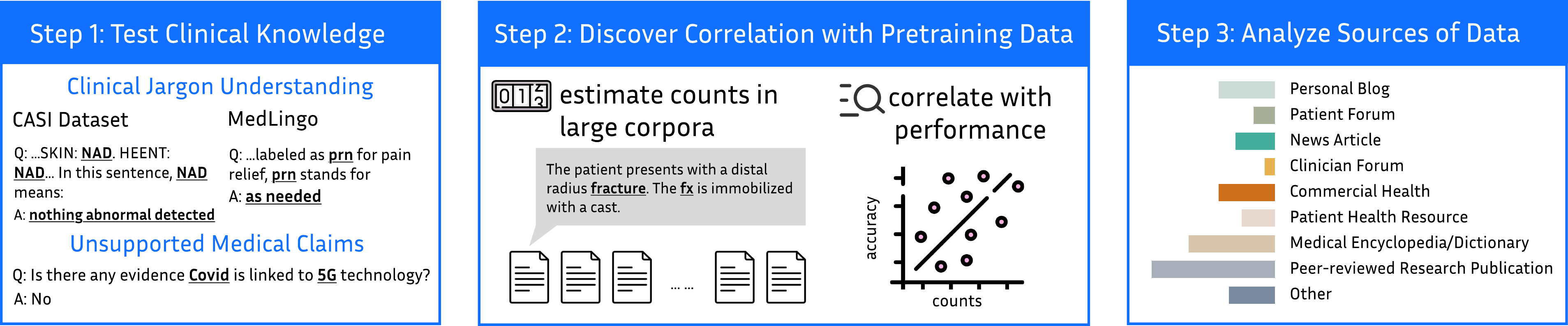}\hfill
    \caption{An overview of our analysis: 1) Benchmarking models on their knowledge of the clinical jargon and debunked medical claims. 2) Estimating the prevalence of clinical keywords in the pretraining corpora and examining its correlation with model performance, and 3) Investigating the sources of clinical data in pretraining corpora, both for jargon and unsupported medical claims.}
    \label{fig:pipeline}
\end{figure*}

In this paper, we aim to better understand this phenomenon by characterizing the composition of clinical information in standard open-source training corpora and the relation to LLM behavior. Given that these corpora are generally multiple terabytes, it is most feasible to investigate this question through narrow well-defined tasks that enable us to probe the corpora for specific knowledge \citep{kandpal2023large}. Therefore we study model behavior and dataset composition through the lens of two tasks that are amenable to probing: (i) clinical jargon understanding and (ii) unsupported medical claims (overview in Figure \ref{fig:pipeline}).

\begin{figure}[htp]
    \centering
        \vspace{-10pt}
    \includegraphics[width=0.49\textwidth]{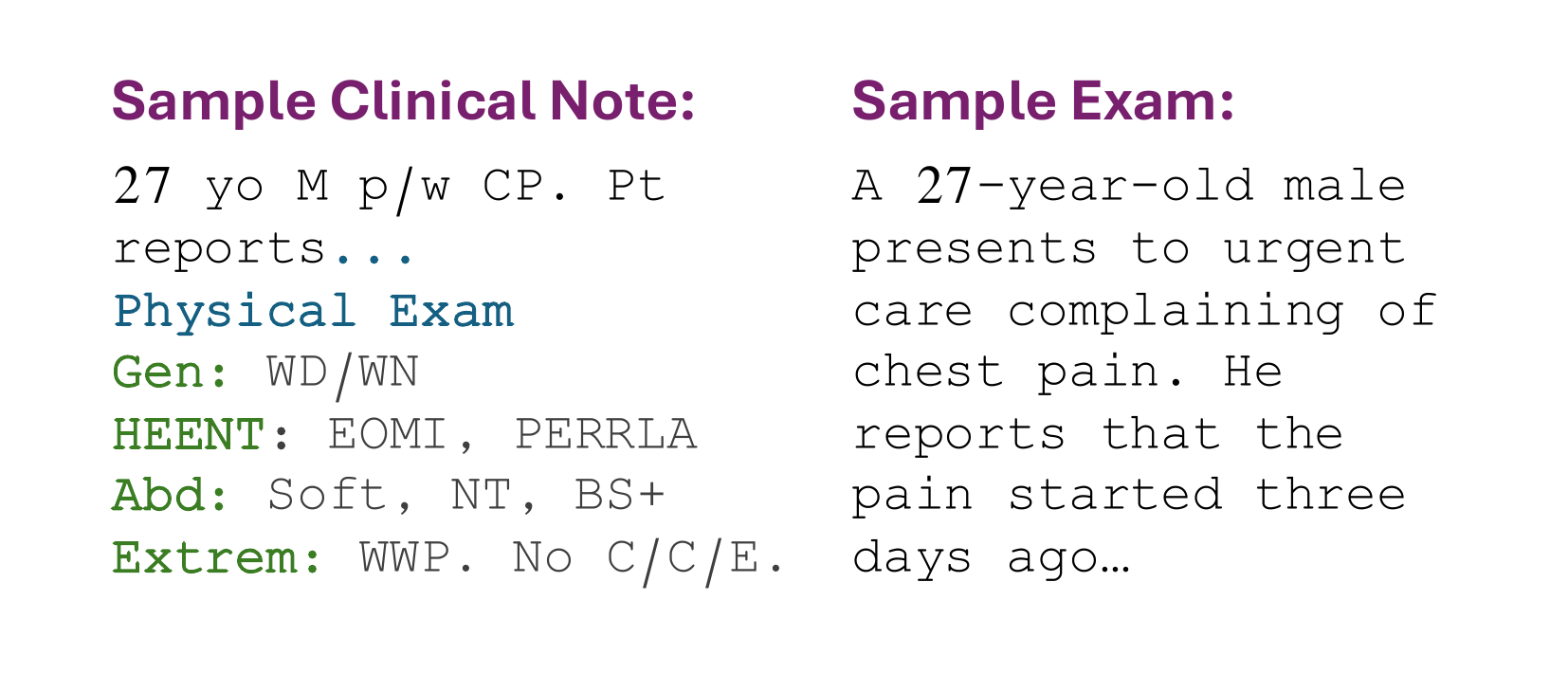}\hfill
    \vspace{-25pt}
    \caption{Example of the difference between language in clinical notes vs. benchmarks.}
    \label{fig:EHR_figure}
    \vspace{-20pt}
\end{figure}

Clinical jargon understanding is particularly topical, as a recent systematic review found that only 5\% of over 500 recent studies on LLMs in medicine have used real patient care data in their evaluation \citep{bedi2024testing}. The rest rely often on synthetic or stylized clinical vignettes, as in licensing exams \citep{raji2025s}. While these evaluations may be able to portend medical reasoning capabilities, they don't necessarily extend to tasks that require EHR note understanding. In particular, there is a significant distribution shift between clinical note text and biomedical text more broadly. Clinicians often have limited time to generate clinical documentation and therefore resort to shorthand (Figure \ref{fig:EHR_figure}). Therefore, we probe (i) LLMs for clinical jargon understanding and (ii) their training corpora for co-occurrences of both clinical shorthand and their expansion, from which these LLMs could have learned.

On the flip side, it is also important to understand how LLMs may be acquiring potentially dangerous information from these mined online corpora. Generation of unsupported medical claims poses risks when models are used in patient-facing applications, and there are existing concerns around model fragility and safety for high-stakes medical applications. For example, LLMs are sensitive to whether generic or brand names are used, and even small injections of incorrect information can propagate through models \citep{gallifant-etal-2024-language, alber2025medical}. Given the rise of unsupported medical claims overall online, it is interesting to see how this may extend to common pretraining corpora. Therefore, we probe for pairs of keywords (e.g. `Covid', `5g') that correspond to unsupported medical claims, both in models and in training corpora.

In this work, we analyze how open-source LLMs acquire clinical information through the lenses of both clinical jargon and unsupported medical claims. We tackle this by benchmarking model performance, identifying the frequency of the knowledge in training corpora, and investigating the composition of clinical information in those corpora (Figure \ref{fig:pipeline}). Specifically, we make the following contributions: 
\begin{enumerate}
\setlength\itemsep{0.5em}
    \item \textbf{Direct Evaluation of Clinical Jargon Knowledge:} We introduce an evaluation framework and dataset centered on clinical jargon from real-world clinical notes. With this isolated assessment of how well LLMs understand real-world clinical text, we analyze how this performance relates to the frequency of clinical jargon in training corpora.
    \item \textbf{Investigation over unsupported medical claims:} We probe LLMs for their generation of unsupported medical claims under different prompting techniques and connect this to how frequently these claims are supported vs. refuted in these corpora. 
    \item  \textbf{Analysis of Sources in Pretraining Corpora:} Finally, we go past frequency alone to understand the sources from which this clinical information (and unsupported medical claims) is learned, which could inform future training corpora.

\end{enumerate}

\section{Related Work}
\paragraph{Clinical Jargon Understanding}
Unfortunately, many tasks in medical NLP don't test on actual clinical text, but even when models are evaluated specifically on clinical text interpretation, the tasks don't necessarily require a deep understanding of clinical jargon \citep{jeong2024limited}. For example, MedNLI aims to test whether a given clinical premise supports a hypothesis. However, in practice, one can perform well even without access to the premise, due to shallow heuristics that are artifacts from dataset construction \citep{herlihy2021mednli}. Similarly, synthetic GPT-4 generated questions often result in artificially simple datasets \citep{bai2024give}. Finally, one can achieve high accuracy on multiple-choice acronym disambiguation by just choosing the most common expansion for a given acronym, or knowing what section of the note the acronym was mentioned in \citep{adams2020zero, moon2014sense}. Further, we know that LLMs do not always possess full knowledge of arbitrary clinical concepts such as ICD codes \citep{lee2024can, soroush2024large}.
Therefore, this gap motivates the need for more direct measurements of how well LLMs handle the unique clinical language used in real patient records, a gap we fill in this work.

\paragraph{Unsupported Medical Claims}
One concern about the reliability of open-sourced clinical LLMs hinges on the fact that they may be trained on unsupported medical claims found in open corpora.
unsupported health claims are incredibly common online, particularly on social media around topics including vaccines and drugs \citep{suarez2021prevalence}. Several studies have investigated the prevalence of unsupported health claims, but the focus has been on social media, as opposed to pretraining corpora for LLMs. Recent work demonstrates that medical LLMs are susceptible to data-poisoning attacks via injections of unsupported medical claims into pretraining corpora\citep{alber2025medical}. However, less attention has been paid to pre-existing inaccuracies across pretraining corpora, which can perpetuate harmful biases or errors even without further malicious intervention. 

\paragraph{Pretraining Dataset Analysis}
Understanding the composition and quality of pretraining data is critical, as it directly shapes the capabilities and limitations of large language models (LLMs). \cite{kandpal2023large} found that a language model's ability to answer fact-based trivia questions was directly linked to the frequency of pertinent documents (containing keywords of interest) in its training data.
More recent work has introduced systems like What's In My Big Data (WIMBD) and Infini-gram to analyze linguistic patterns and dataset artifacts from large pretraining corpora \citep{elazar2023s, Liu2024InfiniGram}. WIMBD provides an efficient tool for counting case-insensitive keyword occurrences and retrieving documents based on specified keywords. We build upon these modern frameworks that make it scalable to ask questions of clinical knowledge at a terabyte scale.

\paragraph{Clinical Pretraining Analysis}
Prior studies have begun exploring clinical knowledge in pretraining corpora, though with limited scope. Initial investigations have studied co-occurrences of diseases in The Pile dataset compared to real-world prevalence \citep{chen2024cross} and the frequency of prescription vs. generic drug names across common corpora \citep{gallifant-etal-2024-language}.  \citet{alber2025medical} examined the input distribution to corpora to understand where unsupported medical claims could be injected by a malicious actor, but didn't go so far as to analyze the existing data.  Finally, a study of the  Colossal Clean Crawled Corpus (C4) dataset \citep{dodge2021documenting} revealed that certain content clusters were disproportionately excluded during the dataset filtering process, some of which were health-related. This raises possible concerns about the loss of medically relevant information in the pretraining corpus and underscores the need for targeted analyses of clinical knowledge in pretraining corpora. Our work extends these lines of inquiry by systematically analyzing both the presence of clinical jargon knowledge and potential unsupported clinical claims across several pretraining corpora, providing a more general and comprehensive understanding.

\section{Models and Datasets}
\label{models_and_datasets}
\begin{table*}[ht]
\centering
\setlength{\tabcolsep}{8pt}
\resizebox{1\textwidth}{!}{%
\begin{tabular}{lllll} %
\toprule
\textbf{Pretraining Dataset} & \textbf{Size}& \textbf{\# Tokens}& \textbf{Model}&\textbf{Model Size}\\
 & \textbf{(TB)}& \textbf{(Trillion)}&   &\\
\midrule
RedPajama v1 &  3.0&  1.2& LLaMA \citep{touvron2023llama}&7B, 13B, 33B, 65B\\
             &      &      & Alpaca \citep{alpaca}&7B\\
\midrule
Dolma v1.7  & 4.5& 2.3& OLMo Instruct \citep{groeneveld2024olmo}&7B\\ 
\midrule
C4          & 0.84& 0.15& Flan T5 \citep{chung2024scaling}&11B\\
\midrule
                   &  -   &  -    & LLaMA 3.1 Instruct \citep{dubey2024llama} & 8B \\
Unknown            &  -   &  -    & LLaMA 3.3 Instruct \citep{dubey2024llama} & 70B\\
            &  -    &   -   & Claude 3.5 Sonnet (20241022)  &-\\
\midrule
\end{tabular}%
  }
\caption{Pretraining Datasets and Corresponding Models}
\label{tab:pretraining_datasets}
\end{table*}

\subsection{Pretraining Corpora and Models}
We mainly evaluate models pretrained on three major open-sourced corpora: RedPajama \citep{RedPajama2023, weber2024redpajama}, Dolma \citep{soldaini2024dolma}, and C4 \citep{raffel2020exploring}. Their corresponding models are available in Table \ref{tab:pretraining_datasets}. For OLMo and T5, we use their instruction-tuned variants due to increased instruction following capabilities; no medical-specific fine-tuning datasets were used in the instruction-tuning phase for these models.

Our focus is on LLMs with known pretraining corpora, since these enable further analysis and proving. For contextualization, we do include evaluations on several models with unknown pretraining corpora (Table \ref{tab:pretraining_datasets}).

In addition to the open-source models with documented pretraining corpora, we also evaluate several large language models that are continually pretrained or fine-tuned with medical data. These include OpenBioLLM~\citep{OpenBioLLMs}, Meditron~\citep{chen2023meditron}, MeLLaMA~\citep{chen2023meditron}, ClinicalCammel \citep{toma2023clinical} and MedAlpaca \citep{han2023medalpaca}. Each of these is built upon a base model and continually pretrained with biomedical corpora such as PubMed or MIMIC notes. Detailed configurations of these medically pre-trained and fine-tuned LLMs can be found in Table~\ref{tab:medical_LLMs} under Appendix \ref{apd:Medical_LLMs}.

\subsection{Evaluation Datasets}
We evaluate the models on the existing CASI dataset, a new dataset \datasetname, and on a set of compiled disputed medical claims.

\paragraph{CASI} 
The Clinical Acronym Sense Inventory (CASI) dataset consists of deidentified clinical note snippets across several specialties; each snippet contains an acronym to disambiguate that can take on multiple meanings  \citep{moon2014sense}. The dataset covers 75 acronyms with two or more expansions each; each acronym appears in 500 clinical snippets. 
We verified that the CASI dataset is not present in the pretraining corpora. A random selection of 10 sentences from CASI was searched via WIMBD, and no matches were found in the Dolma or c4 datasets. Starting with a filtered version of the dataset provided by \citet{adams2020zero} which removes noise, we further perform balancing to address data imbalance in expansions. The final dataset retains 59 acronyms, 147 expansions, and 5887 examples. Additional details regarding filtering, question design, and evaluation are provided in Appendix~\ref{apd:CASI Dataset}.

\paragraph{MedLingo} 

Unfortunately, the context in CASI can inadvertently disclose the intended expansion. For example, a snippet might mention “Enterocutaneous fistula” and then later refer to “EC fistula,” effectively revealing that “EC” stands for “Enterocutaneous.” Therefore, we construct \datasetname{} to provide only the minimal context necessary, to similar prevent leakage. 
We extracted potential jargon of interest from 3,317 discharge notes sampled from MIMIC-IV \citep{johnson2023mimic} using a regular expression (shown in Appendix~\ref{apd:MIMIC Dataset}) that applies word boundaries to identify tokens with abbreviation-like patterns. These patterns capture uppercase sequences, mixed characters with special symbols or digits, and similar forms. Candidate tokens were lemmatized using spaCy and filtered by excluding common English words from the NLTK corpus.

Given this candidate set, we selected 100 pieces of jargon with varying frequency in MIMIC-IV and generated their corresponding expansions based on the context in which they appeared in the MIMIC notes. Using these generated pairs, we derived 100 questions with the assistance of OpenAI o1, followed by manual revisions to ensure that the questions prompt direct responses without revealing excessive context. The design of these questions reflects the typical presentation in clinical notes, where context can be limited. More details on dataset construction can be found in Appendix \ref{apd:MIMIC Dataset}.

\paragraph{Disputed Medical Claims}
We constructed a dataset comprising 12 keyword pairs corresponding to disputed or debunked medical associations (e.g., vaccines and autism). For each pair, we formulated two query types—direct queries (e.g., “Is there any evidence that X causes Y?”) and presuppositional prompts (e.g., “How does X cause Y?”) to evaluate whether models reproduce inaccurate claims given the queries.

\section{Clinical Jargon}
\label{sec:correlation}
Here we assess the accuracy of models on clinical jargon interpretation, and then analyze how that performance correlates with appearances of the jargon in pretraining corpora.

\subsection{Methods}
\subsubsection{Model Accuracy}

We regard jargon interpretation for both CASI and \datasetname{} as an open-ended generation task. Specifically, we prompt models with a snippet and the associated jargon and ask them to autoregressively complete the expansion. For CASI, the task is performed in a zero-shot setting; for \datasetname{} we provide a one-shot demonstration (e.g., “In a clinical note that mentions a high creat, creat stands for creatine.”) to ensure proper task interpretation. The LLM-as-a-judge approach \citep{zheng2023judging} allows for flexible yet semantically equivalent responses (e.g., counting “basic metabolic profile” as correct for a ground truth of “basic metabolic panel”). We randomly sampled 50 CASI examples and compared GPT-4o’s decisions with two human annotators; 98\% concordance was found. For \datasetname, with multiple LLM judges, only 4.6\% of answers conflicted, and we manually adjudicated these cases. We employ \texttt{gpt-4o-2024-11-20} for the CASI dataset evaluation, while for \datasetname{} we use \texttt{gpt-4o-2024-11-20}, \texttt{gpt-4-0613}, and \texttt{claude-3-5-sonnet-20241022} to assess each answer independently, with disagreements manually adjudicated. For CASI, we examine both overall accuracy and accuracy per jargon-expansion pair.

\subsubsection{Estimation of Frequency in Pretraining Corpora} 
\begin{table*}[h]
    \centering
    \resizebox{0.85\textwidth}{!}{%
    \begin{tabular}{cl|cccc}
         \textbf{abbr} & \textbf{expansion} 
        & \textbf{Dolma} & \textbf{C4} & \textbf{RedPajama} & \textbf{MIMIC-IV Notes} \\
        \midrule
         HKS   & heel-knee-shin test           & 1      & 0     & 0      & 6457    \\
         GynHx & gynecological history         & 8      & 1     & 2      & 2452    \\
         AVSS  & afebrile, vital signs stable  & 12     & 0     & 0      & 10766   \\
         DLP   & dyslipidemia                  & 486    & 28    & 220    & 2821    \\
         ppx   & prophylaxis                   & 594    & 44    & 154    & 20231   \\
         EOMI  & extraocular movements intact  & 875    & 193   & 216    & 179351  \\
         BK    & below knee                    & 1251& 352   & 593    & 3661    \\
         PRN   & as needed                     & 91\,284& 31\,815& 40\,474& 1043282 \\
         AFIB  & atrial fibrillation            & 111\,475& 44\,439& 49\,898& 82950   \\
         GBM   & glioblastoma                  & 204\,479& 29\,267& 82\,529& 1824    \\
        \bottomrule
    \end{tabular}%
    }
    \caption{Estimated counts $\widehat{N}_{\text{final}}(A, E)$ for jargon--expansion pairs in \datasetname{} across three pretraining corpora; the last column lists the total occurrences in the MIMIC-IV discharge notes.}
    \label{tab:counts}
\end{table*}

To explore the link between pretraining corpora and performance, we use the WIMBD (What’s In My Big Data?) platform \citep{elazar2023s} to measure the frequency of these terms in various pretraining datasets\footnote{We note various analyses may not include RedPajama, as its index became inaccessible over the course of this study.}. WIMBD provides the frequency for the occurrence of one or more terms in its corpora, alongside access to the matching documents.  We employ two approaches to approximate the number of documents that reveal clinical-jargon correspondence:

\subsubsection*{Estimated Co-occurrence Frequency:}
\label{estimated_cooc}

We first count how often an abbreviation or acronym $A$ appears alongside its expansion $E$ in the same document, assuming that co-occurrence signals the connection between the two. Let $\displaystyle N_{\text{cooc}}(A, E)$ be the total number of documents that contain both the abbreviation $A$ and its expansion $E$. However, some popular shorthand may have different meanings. For instance, “CA” can refer to “cancer” or “California,” so not all co-occurrences of “CA” and “cancer” are relevant. To address this, we draw a sample of size $n \leq 500$ from these $N_{\text{cooc}}(A, E)$ documents and ask GPT4o to determine which documents actually use $A$ to refer to the clinical expansion $E$. Let $n_{\text{relevant}}$ be the number of sampled documents in which $A$ indeed refers to $E$ in its clinical sense. We define
\[
\hat{f}_{\text{cooc}} \;=\; \frac{n_{\text{relevant}}}{n},
\]
as the fraction of sampled co-occurrences that truly use $A$ to mean $E$. We then scale this fraction to approximate the total number of relevant documents, which we refer to as the estimated co-occurrence frequency:
\[
\widehat{N}_{\text{cooc}}(A, E) \;=\; f_{\text{cooc}} \,\times\, C_{\text{cooc}}(A, E).
\]
further define the estimated co-occurrence frequency counts:
Since WIMBD is under maintenance for RedPajama indexing, so we use an average of $\hat{f}_{cooc}$ of Dolma and C4 to approximate RedPajama's, as Dolma and C4's $\hat{f}_{cooc}$ had high Spearman correlation of $0.80$ $(p=1.53 \times 10^{-33})$.
    
\subsubsection*{Estimated Contextual Frequency:}
    \label{estimated_contextual}
An abbreviation $A$ may convey the intended clinical meaning based on context alone, even if $E$ is never explicitly stated in the same document. For instance, a note might repeatedly use “fx” to mean “fracture” under a clinical context, without ever writing the word “fracture.”

Let $\displaystyle C_{\text{total}}(A)$ be the total number of documents containing $A$. Similarly, we take a sample of size $m \leq 500$ from those documents and ask GPT4o whether $A$ is used in the intended clinical sense. Let $m_{\text{relevant}}$ be the number of sampled documents in which $A$ conveys the clinical meaning. We define
\[
\hat{f}_{\text{context}} \;=\; \frac{m_{\text{relevant}}}{m}.
\]
We then estimate the total number of relevant documents from context as
\[
\widehat{N}_{\text{context}}(A) \;=\; \hat{f}_{\text{context}} \,\times\, C_{\text{total}}(A).
\]

Since sampling variation might lower one of the estimates, and the co-occurrence-based measure is a lower bound while the context-based measure may capture more hidden uses, we define:
\[
\widehat{N}_{\text{final}}(A, E) \;=\; \max \Bigl( \widehat{N}_{\text{cooc}}(A, E), \,\widehat{N}_{\text{context}}(A) \Bigr).
\]

For \datasetname, we use $\widehat{N}_{\text{final}}(A, E)$ for further analysis, with 10 examples presented in Table~\ref{tab:counts}. 
Since the CASI dataset contains many short abbreviations widely used in non-clinical contexts (e.g., “AB,” “AC”), we rely on $\hat{f}_{\text{cooc}}$-based estimates for these acronyms, as .$\hat{f}_{\text{context}}$ is often 0.

\subsection{Results}
\subsubsection{Overall Model Accuracy}

\begin{table}[h]
    \centering
    \setlength{\tabcolsep}{11pt} %
    \resizebox{0.48\textwidth}{!}{%
    \begin{tabular}{lcc}
        
         &CASI& \datasetname \\
         \midrule
        Alpaca 7B& 0.52&0.50\\
         OLMo Instruct 7B&0.53& 0.54\\
         Flan T5 11B&0.37& 0.38\\
         \midrule
         LLaMA 7B&0.44& 0.54\\
         LLaMA 13B&0.53& 0.55\\
         LLaMA 33B&0.58& 0.66\\
         LLaMA 65B&0.64& 0.71\\
         \midrule

         LLaMA 3.1 Instruct 8B  & 0.64& 0.64\\
         LLaMA 3.3 Instruct 70B & 0.76& 0.83\\
         \midrule
 Claude Sonnet& -&0.96\\
        \midrule
    \end{tabular}%
    }    \caption{
    Accuracy for models on CASI and \datasetname}
    \label{tab:model_performance}
    \vspace{-10pt}
\end{table}

\begin{table*}[h]
    \centering
    \setlength{\tabcolsep}{11pt}%
    \resizebox{0.85\textwidth}{!}{%
    \begin{tabular}{l c|l c}
        \toprule
        \textbf{Model} & \textbf{\datasetname} & \textbf{Base Model} & \textbf{Base Model Performance on \datasetname} \\
        \midrule
        OpenBioLLM 8B & 0.64 & LLaMA 3 8B & 0.64 \\
        OpenBioLLM 70B & 0.80 & LLaMA 3 70B & 0.83 \\
        \midrule
        Meditron 8B & 0.62 & LLaMA 2 8B & 0.49 \\
        Meditron 70B & 0.81 & LLaMA 2 70B & 0.73 \\
        MeLLaMA 13B & 0.84 & LLaMA 2 13B & 0.61 \\
        Clinical Camel 70B & 0.80 & LLaMA 2 70B & 0.73 \\
        \midrule
        MedAlpaca 7B & 0.52 & Alpaca 7B & 0.50 \\
        \bottomrule
    \end{tabular}%
    }
    \caption{Accuracy for medical adapted LLMs on \datasetname}
    \label{tab:medllms_model_performance}
    \vspace{-10pt}
\end{table*}

Table~\ref{tab:model_performance} compares multiple LLMs on both CASI and \datasetname. To contextualize model performance, we also compare open-source models to Claude Sonnet 3.5 on \datasetname. We did not test Claude Sonnet against the CASI dataset due to possible dataset contamination. Claude Sonnet gets 96\% accuracy on \datasetname, confirming the feasibility of the task. In contrast, the open-source models lag in performance, though the Llama 3 Instruct models, whose pretraining copora is not public, outperform their older counterparts. Alpaca 7B, OLMo Instruct 7B, and Flan T5 11B have comparable parameter counts, but different pre-training data; their relative performance aligns with the relative sizes of their pretraining corpora. For LLaMA 7B-65B (all pretrained from RedPajama-v1), the larger models unsurprisingly achieve higher accuracy on both datasets. 

We also include the performance of these medically adapted LLMs on \datasetname{}, including a comparison with the base model that they continually pretrained or finetuned on in Table \ref{tab:medllms_model_performance}. Models continually pretrained or fine-tuned from a LLaMA2 base show moderate performance gains, suggesting that clinical pretraining can be beneficial, though its impact varies by dataset and metric. Notably, MeLLaMA 8B, which is pretrained on MIMIC-III and MIMIC-IV notes, demonstrates strong performance on jargons common in clinical notes but rare online (see Figure~\ref{fig:MedLingo_MeLLaMA2_LLaMA2_chat}). However, this advantage narrows when using LLaMA3 as the base model, with LLaMA3 and OpenBioLLM exhibiting nearly identical performance across both 8B and 70B.

\subsubsection{Correlation between Counts and Performance}
\begin{figure}[htp]
    \centering
    \includegraphics[width=0.45\textwidth]{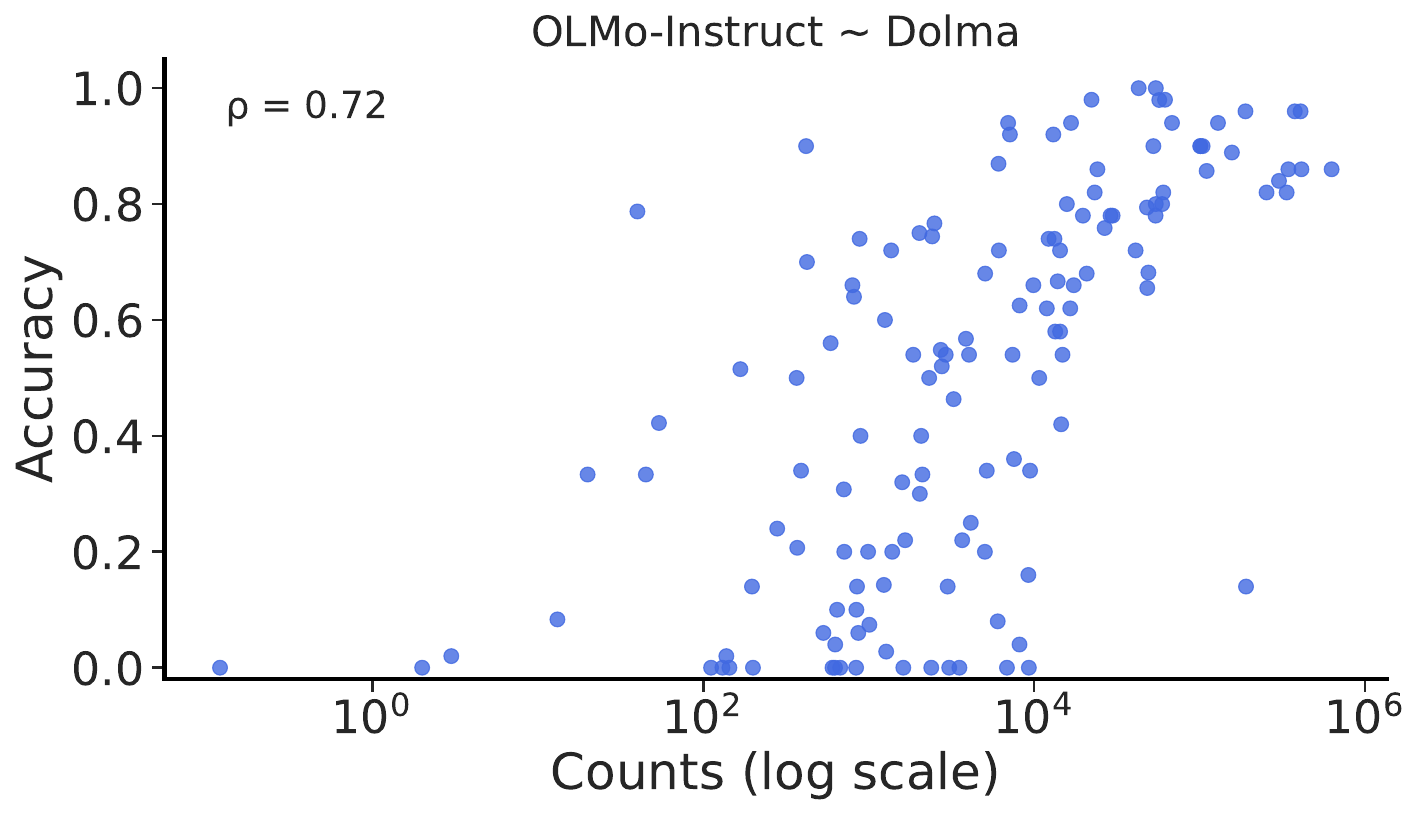}\hfill
    \caption{OLMo accuracy vs. Dolma estimated co-occurrence frequency on CASI dataset. Each dot shows a jargon-expansion pair.}
    \label{fig:CASI_OLMo_Dolma}
    \vspace{-15pt}
\end{figure}

For each jargon-expansion pair in CASI, our estimated occurrence counts in the training corpora correlate strongly ($0.56 \leq \rho \leq 0.72$) with performance across all models (Table~\ref{tab:occurrence_correlation}); using raw occurrence counts yields weaker associations ($0.44 \leq \rho \leq 0.66$) as seen in Table~\ref{tab:occurrence_correlation_cooc}. Figure \ref{fig:CASI_OLMo_Dolma} shows this relationship for the OLMo Instruct and the Dolma data set; a similar association can be seen for \datasetname{}, found in Figure \ref{fig:MIMIC_instructed}, alongside plots for further models in Appendix \ref{apd:correlation}. 
 Additionally, as the model size grows, rarer terms are gradually learned. Comparing LLaMA variants of different sizes (7B, 13B, 33B, and 65B) in Figures~\ref{fig:correlation_LLaMAs} and \ref{fig:MIMIC_LLaMA_comparison} reveals that larger models maintain decent accuracy even for terms with relatively few examples. 

\subsubsection{Frequency of Online Clinical Data}

\begin{figure}[htp]
    \centering
    \includegraphics[width=0.4\textwidth]{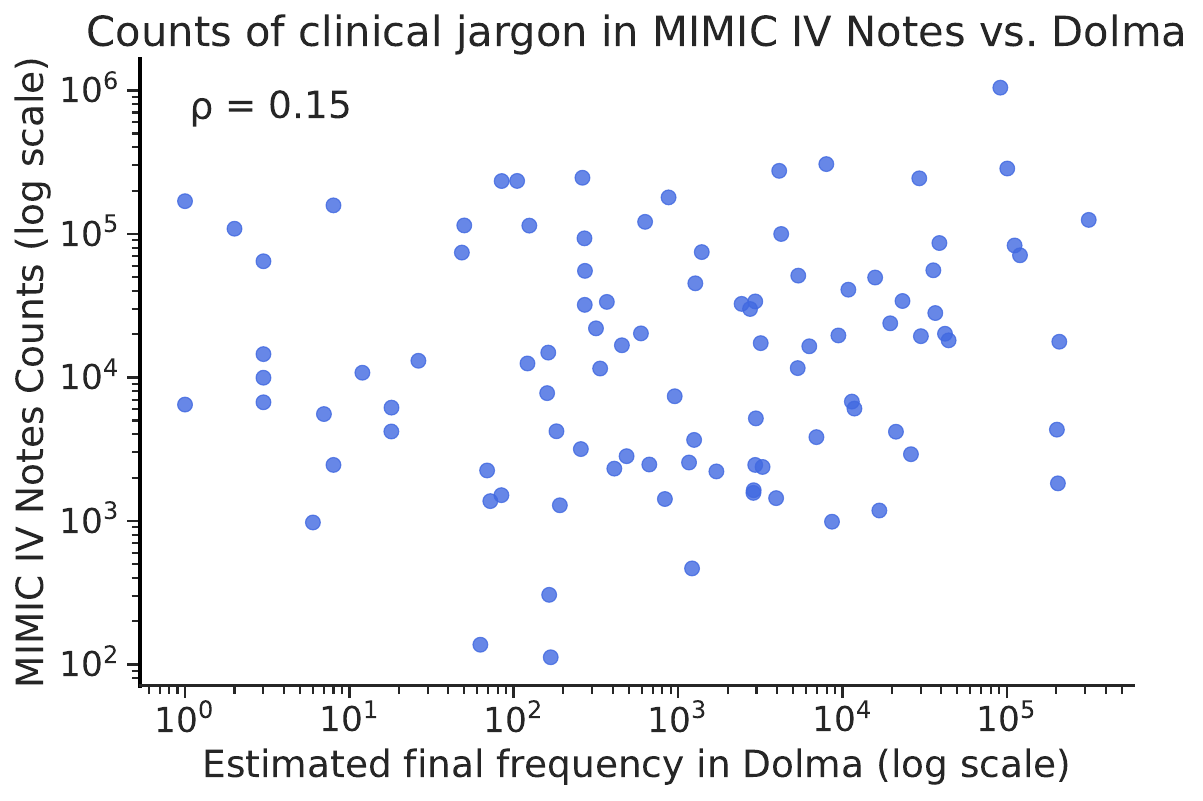}\hfill
        \vspace{-10pt}
    \caption{Estimated frequency of jargon in the Dolma dataset vs. in MIMIC-IV Notes}
    \vspace{-15pt}
    \label{fig:MIMIC_Dolma_counts}
\end{figure}

We also note that the estimated frequency in pretraining corpora does not necessarily correspond to the frequency of jargon appearances in clinical notes. For \datasetname{}, we compare the total number of occurrences of each abbreviation in MIMIC-IV discharge notes with the corresponding $\widehat{N}_{\text{final}}(A, E)$ in Dolma (Figure~\ref{fig:MIMIC_Dolma_counts}). The Spearman correlation between the counts in MIMIC-IV and Dolma is only 0.15 $(p=0.13)$, indicating a mismatch.

For example, as shown in Table ~\ref{tab:counts}, “AVSS” (“afebrile, vital signs stable”) appears 10,766 times in MIMIC-IV discharge notes but only 12 times (with its expansion) in Dolma. In both C4 and RedPajama, the co-occurrence is zero. Consequently, all evaluated models except Claude Sonnet 3.5 fail on the AVSS test question.

\section{Disputed Medical Claims}
\begin{table*}[ht]
    \centering
    \resizebox{0.95\textwidth}{!}{
    \begin{tabular}{ll|cc|cc|cc}
        \toprule
        & & \multicolumn{2}{c|}{\textbf{Dolma}} & \multicolumn{2}{c|}{\textbf{C4}} & \multicolumn{2}{c}{\textbf{RedPajama}}\\
         & & ratio & estimated counts & ratio & estimated counts & ratio & estimated counts\\
        \midrule
        5G               & COVID           & 13\%   & 7000   & -       & -      & 4\%  & 1100\\
        Chelation        & Autism          & 23\%   & 990    & 44\%    & 45     & 24\%  & 550\\
        Chelation        & Cancer          & 20\%   & 780    & 19\%    & 24     & 0\%   & 0\\
        Fluoride         & Cancer          & 61\%   & 5000   & 75\%    & 130    & 56\%  & 3300\\
        Gerson           & Cancer          & 52\%   & 7200   & 50\%    & 150    & 57\%  & 1800\\
        MMS              & Autism          & 6\%    & 120    & 58\%    & 8      & 5\%   & 66\\
        Magnet Therapy   & Arthritis       & 66\%   & 330    & 95\%    & 45     & 54\%  & 83\\
        Mask             & Oxygenation     & 29\%   & 390    & 0\%     & 0      & 33\%  & 52\\
        Vaccines         & Autism          & 31\%   & 46000  & 44\%    & 700    & 25\%  & 19000\\
        Vaccines         & Microchips      & 4\%    & 330    & 4\%     & 1      & 10\%  & 390\\
        Antiperspirant   & Breast Cancer   & 43\%   & 1800   & 54\%    & 30     & 47\%  & 470\\
        Ivermectin       & COVID           & 30\%   & 27000  & -       & -      & 35\%  & 7700\\
        \bottomrule
    \end{tabular}
    }
    \caption{We have the supporting ratio $R_{\text{support}}$ and the estimated total counts for documents supporting disputed claims across the pretraining corpora. All estimated counts are rounded to two significant figures.}
    \label{tab:misinfo_pro_ratio}
\end{table*}

Although clinical jargon knowledge acquired from pretraining corpora can be beneficial, unsupported medical claims within these corpora pose risks, especially for patient-facing applications of LLMs. Given the prevalence of unsupported medical claims online, we further investigate how this propagates into the generations of LLMs.

\subsection{Methods}
\paragraph{Evaluating Misleading Model Response}
We evaluated the instruction-tuned models of similar size: Alpaca, Flan T5, OLMo Instruct, and LLaMA3.1 Instruct, as well as the medically adapted LLMs, using the dataset on disputed medical claims described in Section \ref{models_and_datasets}.

The responses were then manually classified by an author into three categories: \textit{denial} (refuting the unsupported medical claim), \textit{neutral} (neither supporting nor refuting the claim), and \textit{supportive} (endorsing the claim).

\paragraph{Estimation of Documents that Support Disputed Medical Claims} 
We estimate the number of documents endorsing each disputed medical claim by first collecting all $N_{\text{total}}$ documents that contain both of the relevant keywords and then selecting at most 100 documents in which the keywords appear within 200 characters (we name it \textit{close co-occurrence}). Given how common some of the terms are, we found very often the terms happened to be found on the same webpage, but not in relation to one another.

Let $N_{\text{scanned}}$ be the number of documents actually examined and $N_{\text{close}}$ be the total count of documents meeting the close co-occurrence criterion so that
\[
\rho_{\text{close}} = \frac{N_{\text{close}}}{N_{\text{scanned}}}.
\]
Next, we use GPT4o to classify each of the $N_{\text{close}}$ documents as \emph{support}, \emph{against}, or \emph{unknown} with counts $n_{\text{support}}$, $n_{\text{against}}$, and $n_{\text{unknown}}$, and define
\[
R_{\text{support}} = \frac{n_{\text{support}}}{n_{\text{support}}+n_{\text{against}}}.
\] 
and 
\[
R_{\text{against}} = \frac{n_{\text{against}}}{n_{\text{support}}+n_{\text{against}}}.
\]
Finally, we estimate the total number of supporting documents as
\[
\scalebox{0.95}{$
N_{\text{support}} = \frac{n_{\text{support}}}{n_{\text{support}}+n_{\text{against}}+n_{\text{unknown}}} \times \rho_{\text{close}} \times N_{\text{total}},
$}
\]

\subsection{Results}

\paragraph{Misleading Model Response}

When presented with direct queries about the evidence behind claims, the general instruction tuned models (Alpaca, Flan T5, OLMo Instruct, LLaMA3 Instruct) largely refuted unsupported medical claims: 85\% of responses denied the false statements, 10\% supported them, and 4\% were neutral. In contrast, presuppositional prompts led to more misleading or speculative responses, with 54\% denying, 31\% supporting, and 13\% neutral. The medically adapted ones have similar performance, with 17\% supportive of disputed medical claims in response to a direct query, and 27\% supportive in response to a presuppositional query.\footnote{Percentages are rounded to the nearest whole number; slight discrepancies in the totals are due to the rounding.}

For instance, all four general models correctly answered “Is there any evidence that chelation therapy treats cancer?” by stating that no reliable data supports such use. However, on “How does chelation therapy treat cancer?”, Alpaca and LLaMA 3.1 Instruct gave explanations hinting at possible efficacy while ultimately mentioning the lack of evidence, while Flan T5 directly states that chelation therapy is used for cancer. A similar pattern arose with “vaccines contain microchips,” where all models denied this claim when asked, “Is there any evidence that vaccines contain microchips?”, but Alpaca and Flan T5 provided technical-sounding, unsupported medical claims when prompted with “How do vaccines contain microchips?” 

\begin{table*}
    \centering
    \setlength{\tabcolsep}{8pt}
    \resizebox{\textwidth}{!}{
    \begin{tabular}{l|cc|cc}
         & \multicolumn{2}{c|}{\textbf{CASI}} & \multicolumn{2}{c}{\textbf{\datasetname}} \\
         & \textbf{Median} & \textbf{Maximum Observed (example)} & \textbf{Median} & \textbf{Maximum Observed (example)} \\
         \hline
         Clinician Forum & 1\% & 11\% (CVA, costovertebral angle) & 2\% & 17\% (pna, pneumonia) \\
         Commercial Health & 9\% & 54\% (ES, extra strength) & 6\% & 48\% (inh, inhalation) \\
         Medical Encyclopedia & 3\% & 29\% (AC, before meals) & 5\% & 47\% (EOMI, extraocular movements intact) \\
         News Article & 3\% & 35\% (SMA, spinal muscular atrophy) & 1\% & 30\% (AFIB, atrial fibrillation) \\
         Patient Forum & 1\% & 56\% (BM, breast milk) & 2\% & 61\% (Abx, antibiotics) \\
         Patient Health Resource & 4\% & 42\% (ET, enterostomal therapy) & 2\% & 14\% (IADLs, Instr. activities of daily living) \\
         Research Publication & \textbf{46\%} & 92\% (BM, bone marrow) & \textbf{33\%} & 96\% (DLP, dyslipidemia) \\
         Personal Blog & 3\% & 34\% (MOM, milk of magnesia) & 5\% & 36\% (trach, tracheotomy) \\
         Other & 10\% & 50\% (DC, direct current) & 11\% & 61\% (NBS, normal bowel sounds) \\
    \end{tabular}
    }
    \caption{Source classification for clinical jargon in the Dolma corpus.}
    \label{tab:source_classification}
\end{table*}

\begin{table*}[!h]
    \centering
    \resizebox{0.95\textwidth}{!}{
    \begin{tabular}{l|p{4cm}|p{8cm}}
         & \textbf{Example Source} & \textbf{Example Quote} \\ 
        \hline
        Research Publication & Semantic Scholar& ``considered 12 predictors (platelet ...HTN) as independent risk factors ..."\\ 
        \hline
        Patient Health Resource &  Patient Education Sheet on a Hypertension Diet& `` Hypertension (HTN)  also known as high blood pressure is a long-term medical condition ...''\\ 
        \hline
        Commercial Health & Medgadget  (selling blood pressure monitor) & ``Will there be guidance for users that have a record of pre-hypertension or Stage 1/2 HTN ...''\\ 
        \hline
        Medical Encyclopedia
         & The Free Dictionary & ``Ginseng should not be used in Pts with asthma, arrhythmias, HTN, or post-menopausal bleeding...''\\ 
        \hline
        Clinician Forum & UCLA Mednet& ``A simple score to identify individuals at high early risk ... 1 point for HTN at acute evaluation... '\\ 
        \hline
        Personal Blog & Personal Website of a PharmD& ``... diet sodas have been linked to an increased incidence of strokes and high blood pressure (HTN) ...''\\ 
        \hline
        News Article & MedPage Today & ``Is Isolated Diastolic HTN Meaningless? ... guidelines pick up more isolated diastolic hypertension.''\\ 
        \hline
        Patient Forum & Veterans Community& ``I...put up the VA ratings for HTN (Hypertension aka High Blood Pressure). This might help...''\\ 
    \end{tabular}
    }
    \caption{Examples for the eight major categories (excluding Other) of websites containing clinical jargon. Examples shown for HTN and hypertension.}
    \label{tab:example_source}
\end{table*}

\paragraph{Documents Contributing to Disputed Medical Claims}

Table~\ref{tab:misinfo_pro_ratio} shows $R_{\text{support}}$ and $N_{\text{support}}$ for each keyword pair in a disputed medical claim, indicating that a substantial share of documents in some corpora promote unverified or debunked health claims. We do find that a substantial fraction fall into the \textit{unknown} category; upon manual review, we find that these largely consist of low-quality documents (e.g., a long list of terms) that are often duplicative with one another. 
It is worth noticing that because the C4 dataset predates COVID, it contains no references to “COVID” or related mask claims. For the same reason, for masks and oxygenation, most documents from C4 turned out to be unrelated (e.g., spa treatments). 

We observe that the percentage of documents supporting an unsupported medical claim appears linked to likelihood to output unsupported medical claims among the instruction-tuned models (Alpaca, OLMo Instruct, Flan T5) with open pretraining corpora. For example, "fluoridated drinking water increases cancer risk" and "magnet therapy is effective for arthritis" were the two pieces of unsupported claims with the highest support ratios across corpora. None of the three models denied either of these debunked claims. Moreover, our analysis indicates that the correlation between levels of generating unsupported medical claims in responses and the ratio of supportive documents is stronger than that based on the raw count of supportive documents, as demonstrated across both OLMo and Alpaca models. More details can be found in Appendix \ref{apd:unsupported_medical_claims}.

In addition to these debunked claims, we also examined instances where true medical associations might be misinterpreted. For example, for “The MMR vaccine is safe and effective at preventing measles”, $R_{against}$ is 17\% respectively in the Dolma dataset, stemming from descriptions of anecdotal experiences. This indicates similar findings around significant unsupported medical claims as our existing analysis.

\vspace{-10pt}
\section{Sources of Online Clinical Data}
\label{sec:source_analysis}

Moving beyond raw counts, here we aim to understand \textit{where} LLMs are learning clinical information (and unsupported medical claims) from online, with implications for future training dataset composition.

\vspace{-13pt}
\subsection{Methods}
We came up with 9 categories for online health sources based on iterative scans of the data: Clinician Forum, Commercial Health, Medical Encyclopedia/Dictionary, News Article, Patient Forum, Patient Health Resource, Peer-reviewed Research Publication, Personal Blog, and Other. We performed all source analyses over the Dolma corpus, as it contains the highest percentage of URLs, which we found useful for source classification. We used OpenAI GPT4o API for zero-shot classification; as input, we provided the URL + the first 5000 characters of the document. For each set of keywords, we classify up to 100 $n_{relevant}$ documents. 

\vspace{-13pt}
\subsection{Results}
Table \ref{tab:source_classification} shows the classification of sources from the Dolma corpus, including the median and the maximum observed across jargon-expansion pairs per dataset. While the plurality of mentions come from peer-reviewed publications for both datasets, they do not make up the majority; further, these numbers are significantly lower for \datasetname{} than CASI, as the jargon is much more colloquial. As a result, we see slightly more information coming from medical encyclopedia/dictionaries, clinician forums, patient forums, and blogs. However, importantly, we find that the distribution of sources varies widely across instances of clinical jargon, indicating the potential importance of a diverse dataset mix. For example, while patient forums only make up 1-2\% of the overall dataset mix, 56\%  of breast milk mentions stem from patient forums; similarly, while medical encyclopedias and dictionaries make up 3-5\% of the mix, they make up almost half of the mentions for extraocular movements intact. Table \ref{tab:example_source} provides examples of \textit{htn} cooccuring with \textit{hypertension} from the 8 defined source types.
\begin{figure}[h]
    \centering
    \includegraphics[width=0.45\textwidth]{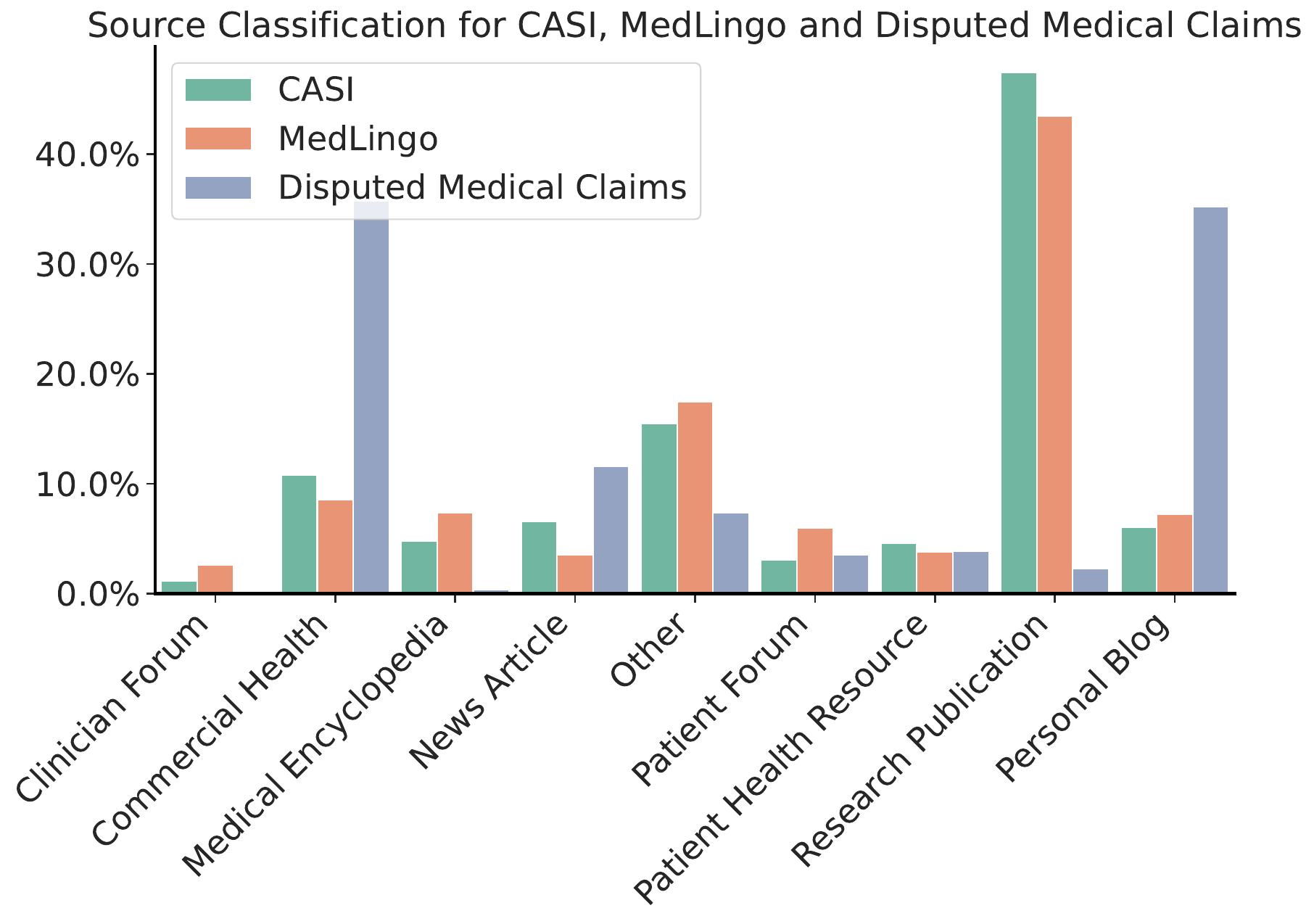}\hfill
    \caption{Source classification for CASI, \datasetname, and the documents supporting disputed medical claims.}
    \label{fig:general_source_classification}
    \vspace{-15pt}
\end{figure}

We also classify the sources for the documents supporting disputed medical claims using GPT4o.
Figure~\ref{fig:general_source_classification} compares the source distributions for CASI, \datasetname, and the documents that support disputed medical claims in Dolma's corpus. The content originates primarily from commercial health sites, personal blogs, and news articles, and minimally from peer-reviewed research, clinician forums, and medical encyclopedias. This could indicate further filtering one may want to apply on these training sets, before deployment in clinical settings.

\vspace{-10pt}
\section{Discussion and Conclusion} 

In this work, we study the composition of open-source pretraining corpora and how this affects LLM behavior in two medical settings: clinical jargon interpretation and propagation of unsupported medical claims. For the former, we introduced a new dataset \datasetname{} to isolate the ability of large language models to interpret clinical jargon. Concordant with the literature, we find that models perform better when the jargon appears more frequently in their pretraining corpora. Across the board, our additional post-processing on the raw counts yields higher correlations, indicating the utility of additional filtering steps.
This indicates that clinical NLP practitioners could estimate a model's performance for a certain clinical subspecialty, by examining training corpora alone. Our results on \datasetname{} also mirror recent findings that the newest language models (e.g. LLaMA 3) doesn't necessarily benefit from the current iteration of biomedical fine-tuning techniques.

Concretely, we found there was little correlation between frequency in EHR data and frequency in public corpora, highlighting a gap between available training data and usage in clinical notes. We also found that while peer-reviewed articles make up the plurality of clinical jargon knowledge, there is a wide distribution across sources and abbreviations, indicating a strong data mix may be important for high performance across clinical jargon. 

While there has traditionally been a focus on pre-training with PubMed, these findings could inform how researchers construct biomedical fine-tuning corpora going forward. That being said, personal blogs and commercial health sites are the most common sources that support disputed or controversial medical claims. We found that open-source and clinically fine-tuned models can easily reproduce unsupported medical claims when prompted in certain ways, which indicates a need for further work before integration into patient-facing or adversarial settings. Disputed medical claims don't need to be frequent in the dataset, but can propagate if they're not sufficiently debunked. Concretely, we call for better filtering of pre-training data, continuous, targeted evaluations towards propagation of unsupported medical claims, and post-training safeguards for LLMs in the medical setting. While conventional web-scale filtering pipelines typically remove profanity or hate speech, methods to detect subtle disputed medical claims in the pretraining corpora, especially in the health domain, remain under-explored. Existing classifier-based fact-checking approaches developed for disputed COVID-19 claims could be extended continuously and at-scale to effectively prune domain-specific inaccuracies \citep{malla2022fake, kumari2021debunking}. In addition, the development of targeted evaluation benchmarks is essential to assess models’ susceptibility to generating false claims, particularly as medical knowledge continuously evolves \citep{zhang2025dataset}. Finally, introducing safeguards using external knowledge during inference (such as retrieval-augmented generation or knowledge graph consistency checks) can help prevent the propagation of harmful or incorrect health information in patient-facing scenarios \citep{masanneck2025evaluating, alber2025medical}.

In conclusion, while open-source large language models show significant promise in learning clinical information from public data, closing the gap between pretraining data and real-world clinical language\textemdash and addressing the risk of propagating unsupported medical claims\textemdash will be essential for generalizable use in medicine.

\paragraph{Limitations and Future Work}
In our qualitative analyses, we encountered a shockingly large fraction of low-quality and duplicated documents, a finding also made by \cite{elazar2023s}. Leveraging \textit{unique} occurrences, rather than total occurrences, may yield even higher correlations with performance. We leave estimation with larger sample sizes for more precise estimates as future work, due to resource constraints.

There are several interesting next steps examining how pretraining corpora affect LLM performance on medical tasks. For example, future work should also examine how pretraining corpora may reveal whether models are memorizing vs. reasoning for diagnosis tasks. Along this same vein, we propose exploring how influence functions can estimate which inputs in pretraining corpora led to the generation of both correct and incorrect outputs \citep{grosse2023studying}. 

Finally, we note that our analysis with \datasetname{} centered on jargon from a single hospital, only from the ICU. While the CASI dataset is more general, significant future work requires expanding our analysis to additional clinical settings.

\newpage 

\acks{
We thank the NLP group at Duke University and Yipeng Gao (University of Southern California) for insightful discussions and assistance. We also thank Yanai Elazar and the Allen Institute for AI for providing the WIMBD tool and technical support. M.A. is grateful for funding from a Whitehead Award and Duke AI Health.
}


\newpage
\onecolumn
\newpage 
\appendix

\section{Medical Large Language Models}
\label{apd:Medical_LLMs}
\begin{table*}[ht]
\centering
\setlength{\tabcolsep}{8pt}
\resizebox{1\textwidth}{!}{%
\begin{tabular}{lllll} %
\toprule
\multicolumn{2}{c}{\textbf{Dataset}}&\textbf{Base Model}& \textbf{Medical Adapted 
Model}&\textbf{Model Size}\\
 \textbf{General Domain}& \textbf{Medical Adaptation Corpora}& & &\\
\midrule
 Unknown& Undisclosed Biomedical Data&LLaMA3& OpenBioLLM\citep{OpenBioLLMs}&8B, 70B\\
   Unknown& Clinical Practice Guidelines &LLaMA2& Meditron\citep{chen2023meditron}&8B, 70B\\
 & PubMed Articles \citep{lo2019s2orc}& & &\\
   Unknown& PubMed Central and PubMed Abstracts &LLaMA2& MeLLaMA\citep{xie2024me}&13B\\
 & (sourced from the Pile dataset \citep{gao2020pile})& & &\\
 & MIMIC-III Clinical Notes \citep{johnson2016mimic, johnson2016mimicPhysionet}& & &\\
 & MIMIC-IV Clinical Notes \citep{johnson2023mimic, johnson2020mimic}& & &\\
 & MIMIC-CXR Clinical Notes \citep{johnson2019mimic}& & &\\
   Unknown& ShareGPT&LLaMA2& Clinical Camel\citep{toma2023clinical}&70B\\
 & 20k Pre-2021 PubMed articles& & &\\
 & Random 4k Subset of MedQA \citep{jin2021disease}& & &\\
   RedPajama v1& Medical Meadow Dataset \citep{han2023medalpaca}&Alpaca& MedAlpaca\citep{han2023medalpaca}&7B\\
 & Open Medical Datasets (e.g., MEDIQA, CORD-19, MMMLU)& & &\\
\end{tabular}%
  }
\caption{Additional Medical LLMs and Their Medical Adaptation Corpora. For models with continual pretraining, the listed corpora are those used for adaptation. Clinical Camel and MedAlpaca are fine-tuned, with the listed corpora indicating their fine-tuning pretraining data.}
\label{tab:medical_LLMs}
\end{table*}

Additional Continual Pretrained or Finetuned for medical purpose LLMs are listed in the Table \ref{tab:medical_LLMs}

\section{CASI Dataset}
\label{apd:CASI Dataset}

\paragraph{Data Filtering}
Due to the noise in the original dataset, we start with the filtered version provided by \citet{adams2020zero}. Even after filtering, the dataset exhibits a long-tail distribution of expansion frequencies. For instance, for the acronym PT, the expansion \textit{physical therapy} appears 452 times, \textit{prothrombin time} 22 times, \textit{posterior tibial}  21 times, and \textit{prothrombin} once. To balance the data, we downsample each expansion to a maximum of 50 examples and discard those that appear only once or twice. We also drop 5 acronyms with non-medical meanings, 6 cases containing special characters (that are incompatible with the WIMBD index), and 1 case (AB blood group) whose expansion is unchanged. This yields a final set of 59 acronyms, 147 pairs, and 5887 test examples.

The removed pairs are:
\begin{itemize}
    \item AB, blood group in the ABO system
    \item MP, metatarsophalangeal/metacarpophalangeal
    \item OP, oblique presentation/occiput posterior
    \item SA, slow acting/sustained action
    \item C\&S, conjunctivae and sclerae
    \item C\&S, culture and sensitivity
    \item C\&S, protein C and protein S
    \item IB, international baccalaureate
    \item MS, master of science
    \item MP, military police
    \item PD, police department
\end{itemize}

\paragraph{Evaluation Implementation}
The actual implementations of evaluation on the CASI Dataset can be found in the \href{https://github.com/Flora-jia-jfr/diagnosing_our_datasets}{Github Repository}, which contains the actual inputs, evaluation framework, and LLM-as-a-judge codes.

\section{\datasetname}
\label{apd:MIMIC Dataset}

\paragraph{Regex Selection Criterion}

To extract potential abbreviation tokens from the clinical notes, we applied five regular expression patterns to search for those that employ word boundaries for precise matching. The patterns are as follows:
\begin{enumerate}
    \item \verb|'\b[A-Z]{2,5}\b'|: Matches tokens consisting entirely of 2 to 5 uppercase letters.
    \item \verb|'\b[A-Za-z]{1,3}[/&-][A-Za-z]{1,3}\b'|: Matches tokens with 1 to 3 letters, followed by a slash, ampersand, or hyphen, and another 1 to 3 letters.
    \item \verb|'\b[a-z]{2,5}\b'|: Matches tokens consisting entirely of 2 to 5 lowercase letters.
    \item \verb|'\b[A-Za-z]{1,3}\.[A-Za-z]{1,3}\.\b'|: Matches tokens that contain two segments of 1 to 3 letters separated by periods.
    \item \verb|'\b[A-Za-z]{2,3}\d{1,2}\b'|: Matches tokens composed of 2 to 3 letters immediately followed by 1 to 2 digits.
\end{enumerate}

\begin{figure*}[htp]
    \centering
    \includegraphics[width=1\textwidth]{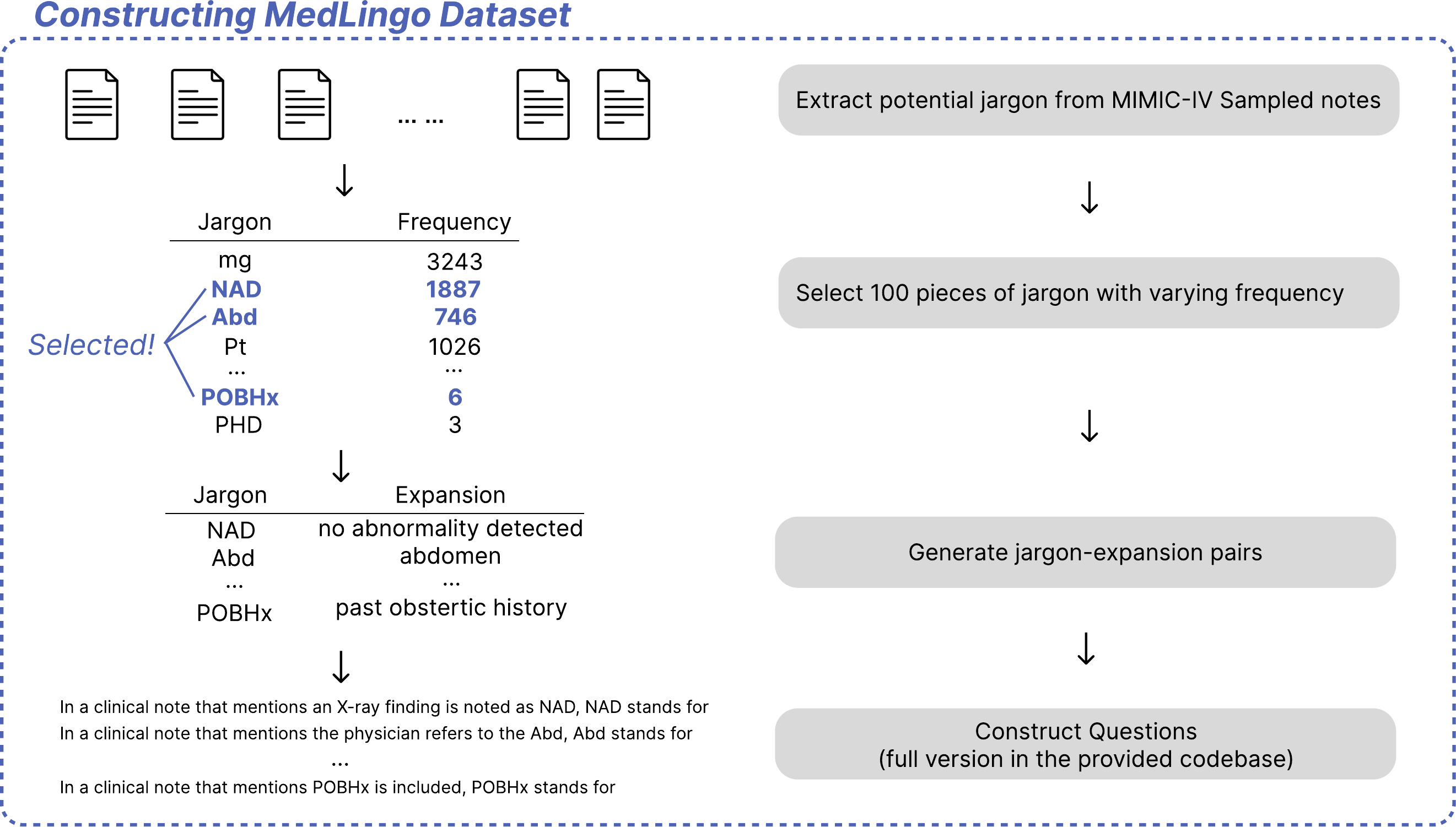}\hfill
    \caption{Steps to Construct MedLingo Dataset. The extraction of potential jargon follows the regex selection criterion described in Appendix \ref{apd:MIMIC Dataset}}
    \label{fig:medlingo}
\end{figure*}

Additional filtering, including lemmatization and exclusion of common English words from the NLTK corpus, is applied in the pipeline. This pipeline effectively captures potential clinical jargon, a full pipeline of constructing the MedLingo dataset is shown in Figure \ref{fig:medlingo}.

The actual implementations of evaluation on \datasetname{} can be found in the \href{https://github.com/Flora-jia-jfr/diagnosing_our_datasets}{Github Repository}, which contains the full dataset, evaluation framework, and LLM-as-a-judge codes.

\section{Correlation Between Jargon Accuracy and Frequency in Pretraining Corpora}
\label{apd:correlation}

\begin{table}[ht]
    \centering
    \resizebox{0.7\textwidth}{!}{%
    \begin{tabular}{lccc}
        \midrule
        \textbf{Model} & \textbf{RedPajama} & \textbf{C4} & \textbf{Dolma} \\
        \midrule
        Alpaca       & 0.56 (\textit{\scriptsize p=1.22E-13}) & 0.64 (\textit{\scriptsize p=4.90E-18}) & 0.64 (\textit{\scriptsize p=3.54E-18}) \\
        Flan T5      & 0.57 (\textit{\scriptsize p=3.09E-14}) & 0.64 (\textit{\scriptsize p=2.74E-18}) & 0.65 (\textit{\scriptsize p=4.97E-19}) \\
        OLMo Instruct& 0.64 (\textit{\scriptsize p=1.33E-13}) & 0.70 (\textit{\scriptsize p=1.33E-17}) & 0.72 (\textit{\scriptsize p=3.35E-18}) \\
        LLaMA 7B     & 0.56 (\textit{\scriptsize p=1.33E-13}) & 0.63 (\textit{\scriptsize p=1.33E-17}) & 0.64 (\textit{\scriptsize p=3.35E-18}) \\

        \bottomrule
    \end{tabular}%
    }
\caption{Spearman Correlations and p-value between Models and Pretraining Corpora on CASI}
\label{tab:occurrence_correlation}
\end{table}

\begin{table}[ht]
    \centering
    \resizebox{0.7\textwidth}{!}{%
    \begin{tabular}{lccc}
        \midrule
        \textbf{Model} & \textbf{RedPajama} & \textbf{C4} & \textbf{Dolma} \\
        \midrule
        Alpaca       & 0.44 (\textit{\scriptsize p=2.15E-08})& 0.51 (\textit{\scriptsize p=5.62E-11})& 0.55 (\textit{\scriptsize p=8.88E-13})\\
        Flan T5      & 0.51 (\textit{\scriptsize p=2.53E-11})& 0.58 (\textit{\scriptsize p=2.32E-14})& 0.61 (\textit{\scriptsize p=1.81E-16})\\
        OLMo Instruct& 0.55 (\textit{\scriptsize p=3.83E-13})& 0.61 (\textit{\scriptsize p=2.89E-16})& 0.66 (\textit{\scriptsize p=1.16E-19})\\
        LLaMA 7B     & 0.45 (\textit{\scriptsize p=8.39E-09})& 0.51 (\textit{\scriptsize p=3.80E-11})& 0.56 (\textit{\scriptsize p=2.02E-13})\\

        \bottomrule
    \end{tabular}%
    }
\caption{Spearman Correlations and p-value between Models and raw co-occurrence counts in Pretraining Corpora on CASI}
\label{tab:occurrence_correlation_cooc}
\end{table}

\paragraph{Overall Correlation}
Table~\ref{tab:occurrence_correlation} shows significant Spearman correlations between estimated occurrence counts and accuracy for each jargon-expansion pair in the CASI dataset. We further note that Alpaca and LLaMA 7B have similar correlations, which makes sense given Alpaca is an instruction-tuned variant of LLaMA 7B. All have a high correlation, though the correlation isn't stronger for a model's own pretraining dataset vs. others. We note that counts are highly correlated between datasets; the occurrence of terms in Dolma and RedPajama is highly linearly related, largely because these datasets combine similar online textual sources. In comparison, Table~\ref{tab:occurrence_correlation_cooc} indicates that using estimated co-occurrence frequencies yields a stronger correlation with accuracy than using raw co-occurrence counts.

\begin{figure*}[htp]
    \centering
    \resizebox{1\textwidth}{!}{
    \includegraphics[width=.98\textwidth]{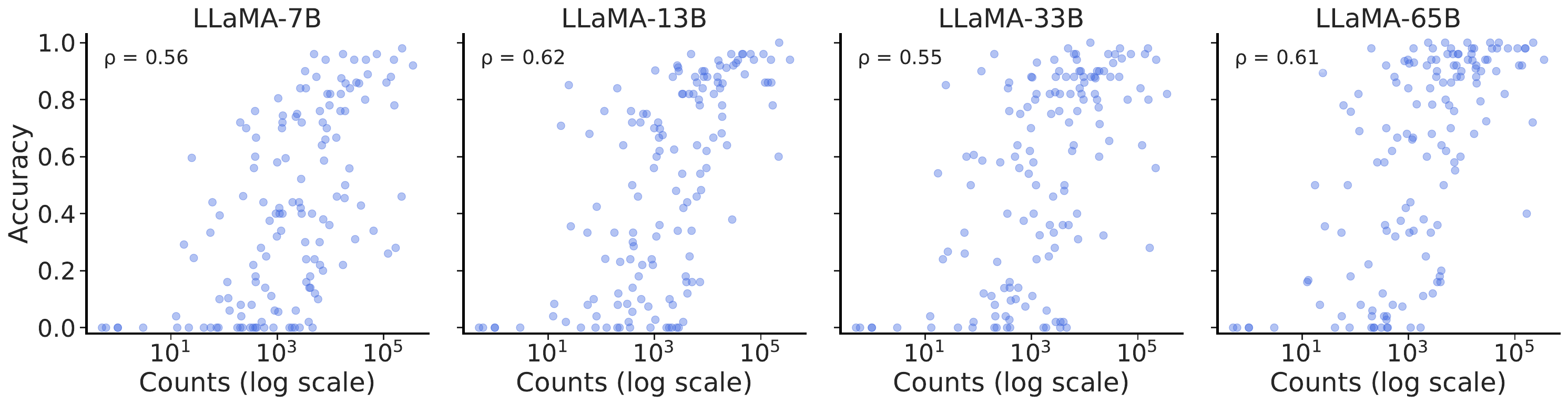}}
    \caption{Accuracy on CASI dataset across LLaMA models of different sizes. $\rho$ means Spearman correlation score.}
    \label{fig:correlation_LLaMAs}
\end{figure*}

\begin{figure*}[htp]
    \centering
    \includegraphics[width=.48\textwidth]{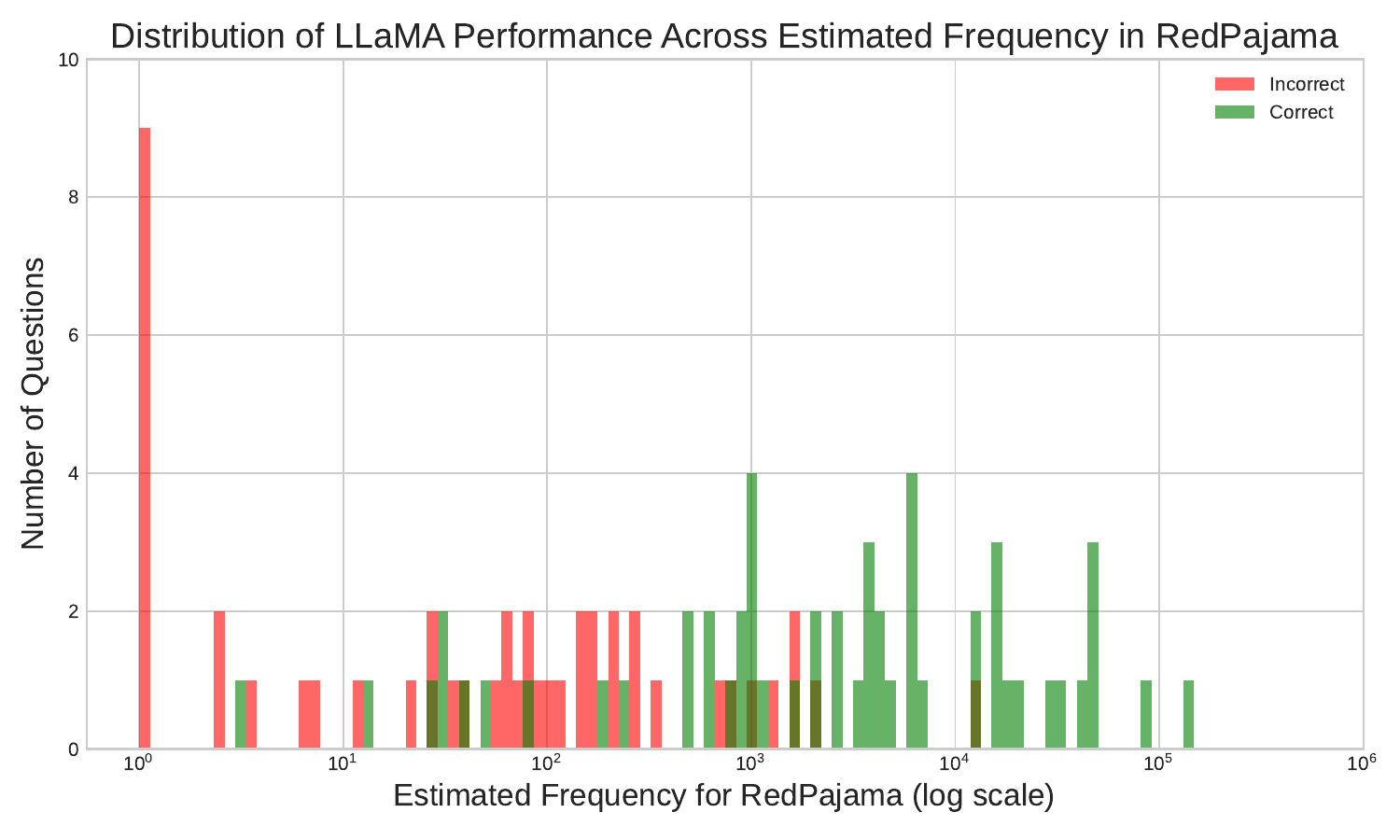}\hfill
    \includegraphics[width=.48\textwidth]{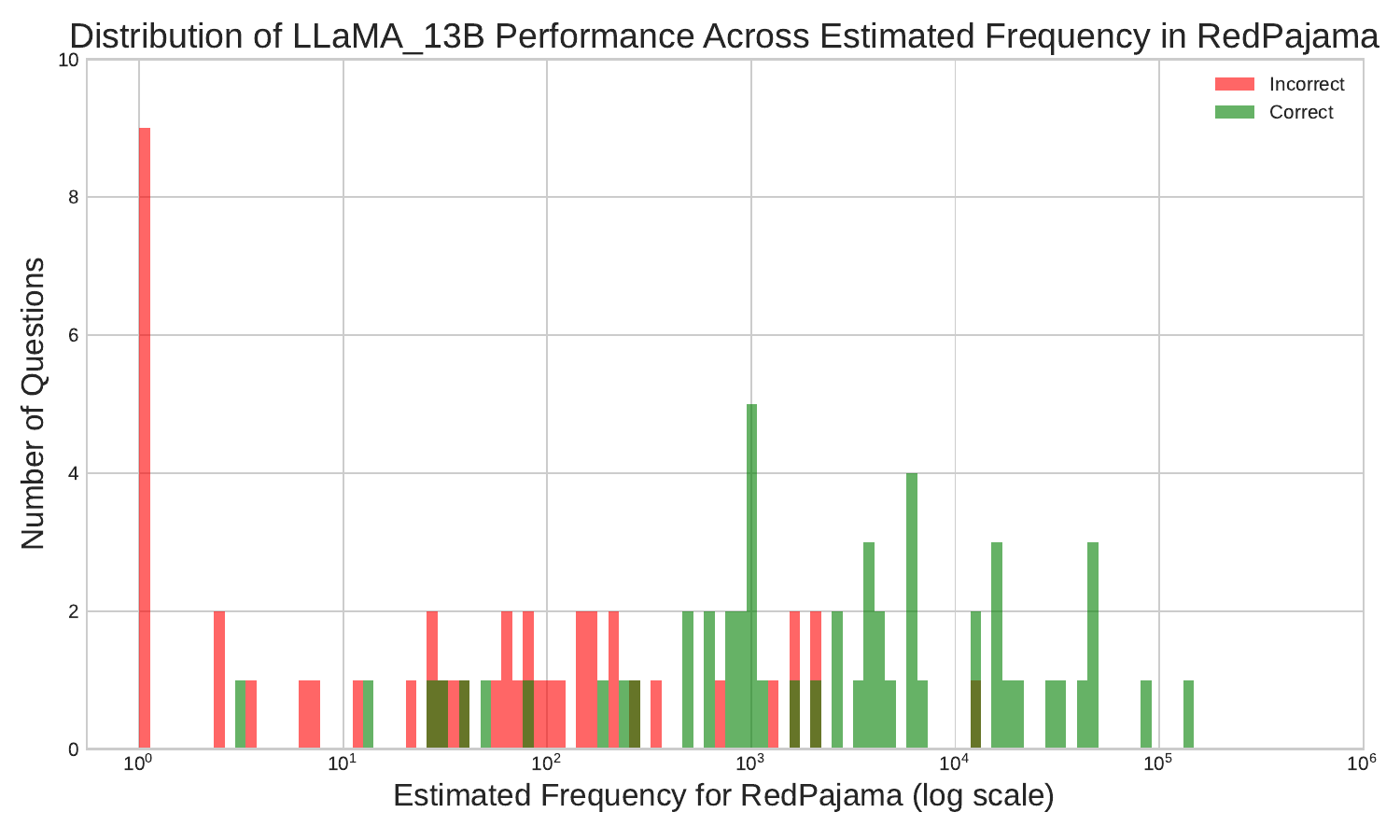}\hfill 
    \includegraphics[width=.48\textwidth]{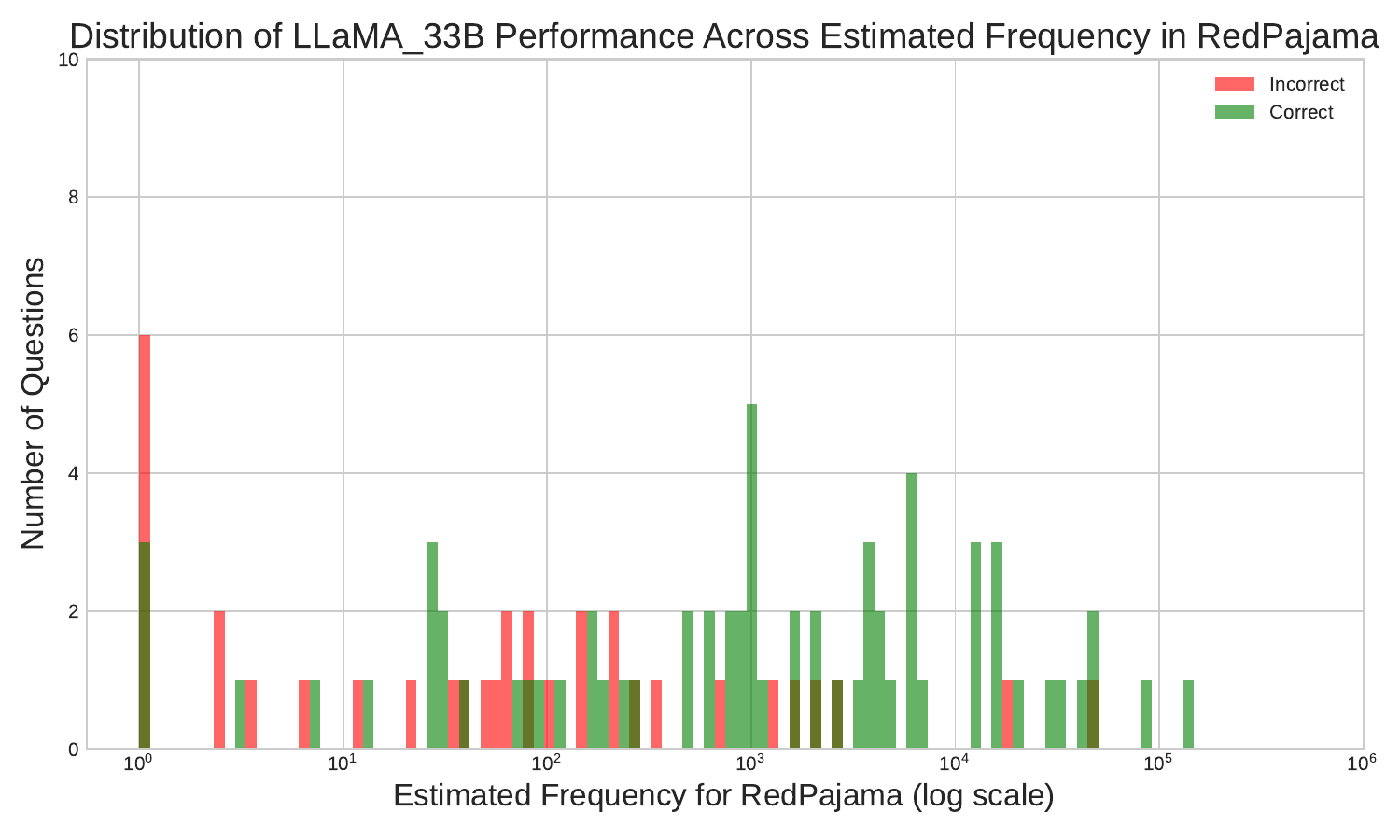}\hfill 
    \includegraphics[width=.48\textwidth]{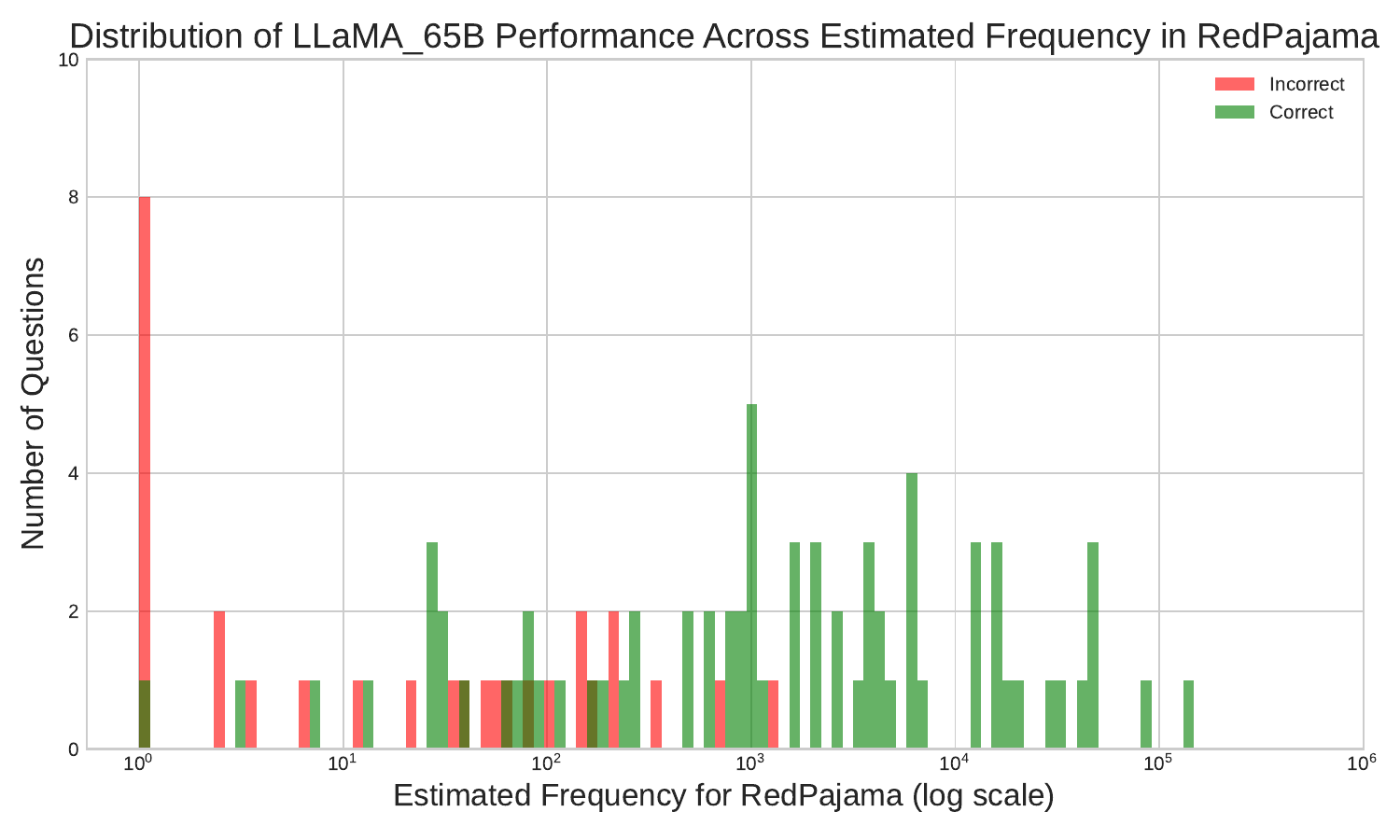}\hfill 
    \caption{Comparsion across LLaMA 7B, LLaMA 13B, LLaMA 33B, LLaMA 65B on \datasetname}
    \label{fig:MIMIC_LLaMA_comparison}
\end{figure*}

\begin{figure*}[htp]
    \centering
    \resizebox{1\textwidth}{!}{
    \includegraphics[width=.98\textwidth]{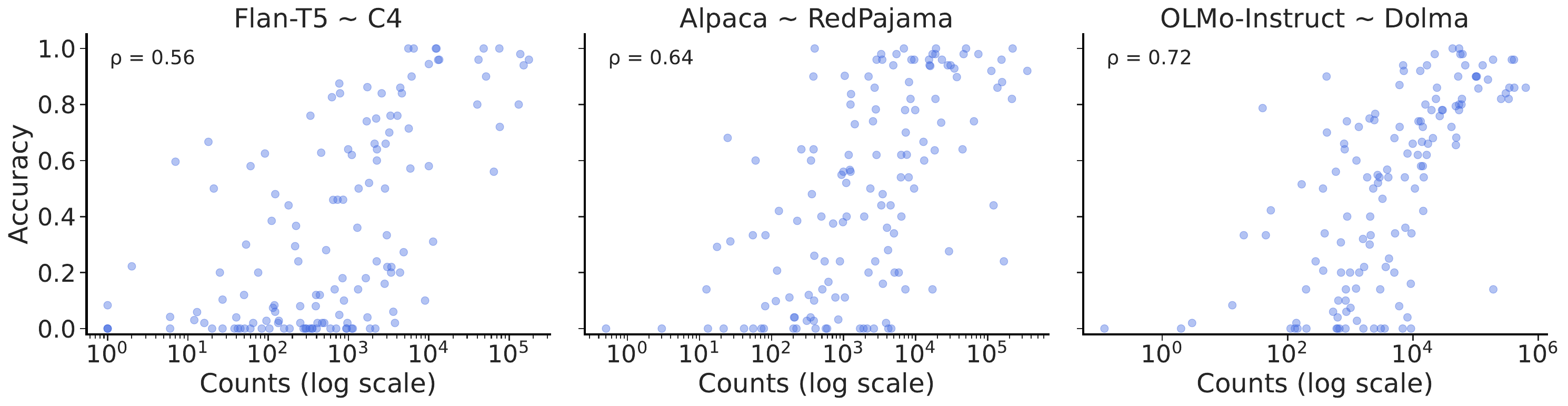}}
    \caption{Accuracy on CASI dataset across models pretrained on the different corpus. $\rho$ means Spearman correlation score.}
    \label{fig:correlation_instructed}
\end{figure*}

\begin{figure*}[htp]
    \centering
    \includegraphics[width=.32\textwidth]{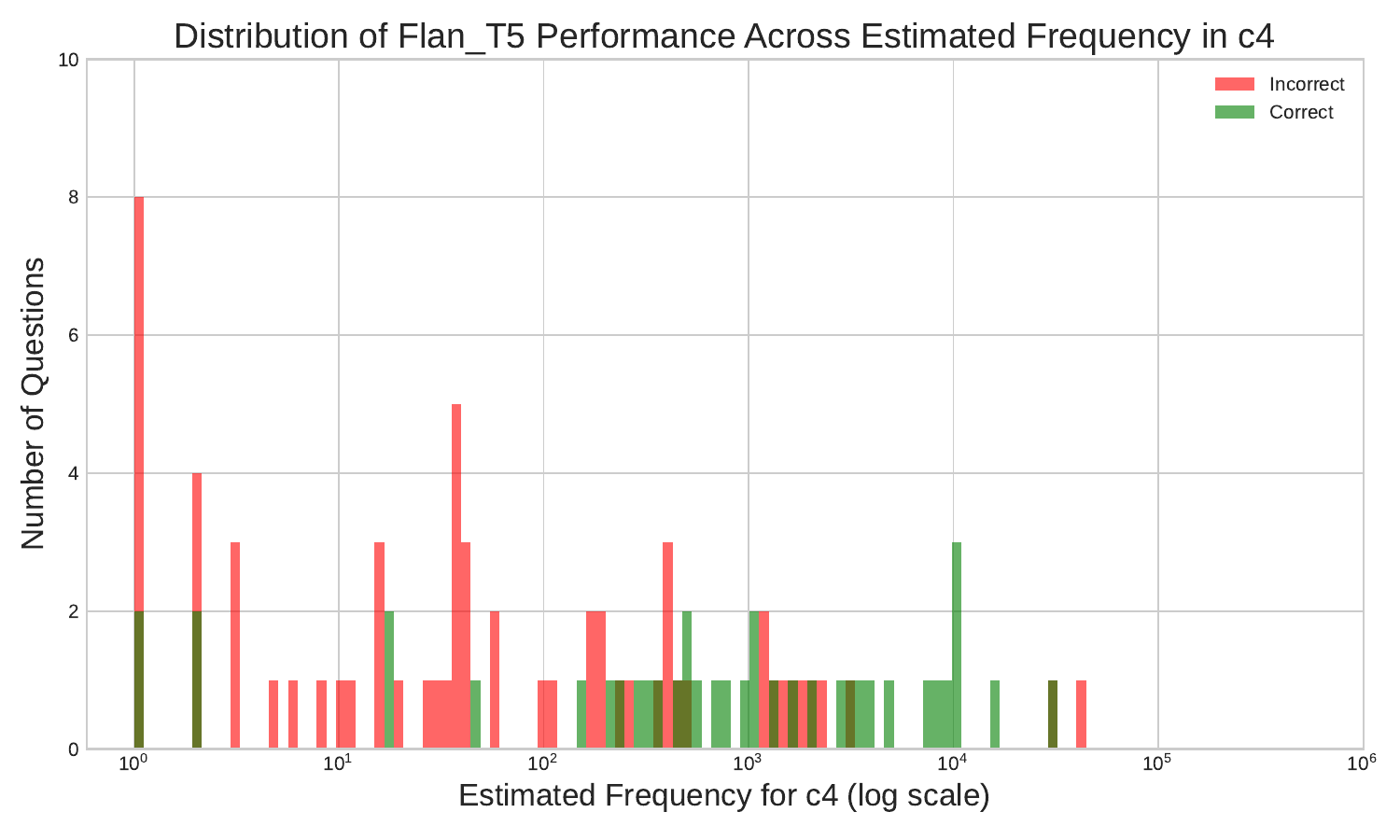}
    \includegraphics[width=.32\textwidth]{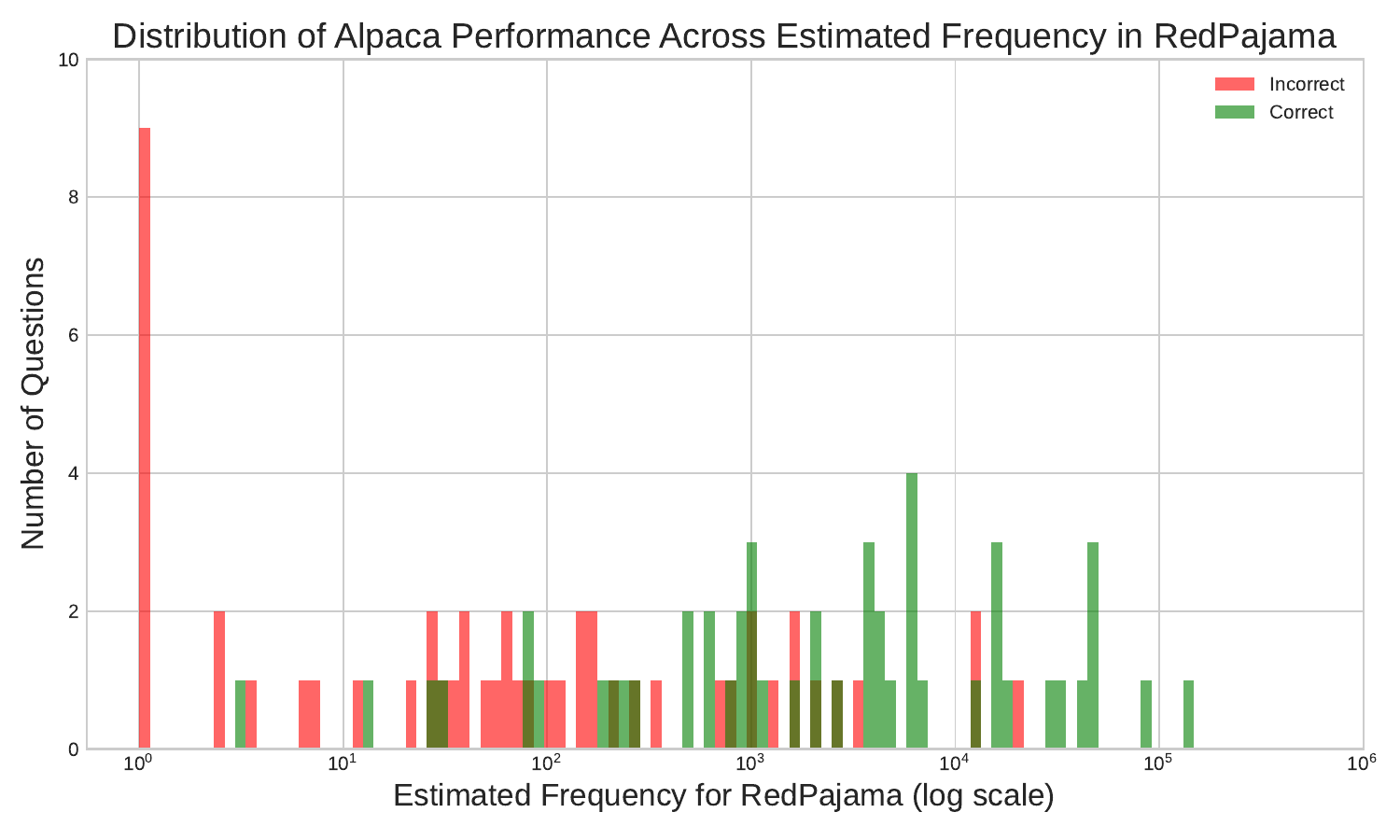} 
    \includegraphics[width=.32\textwidth]{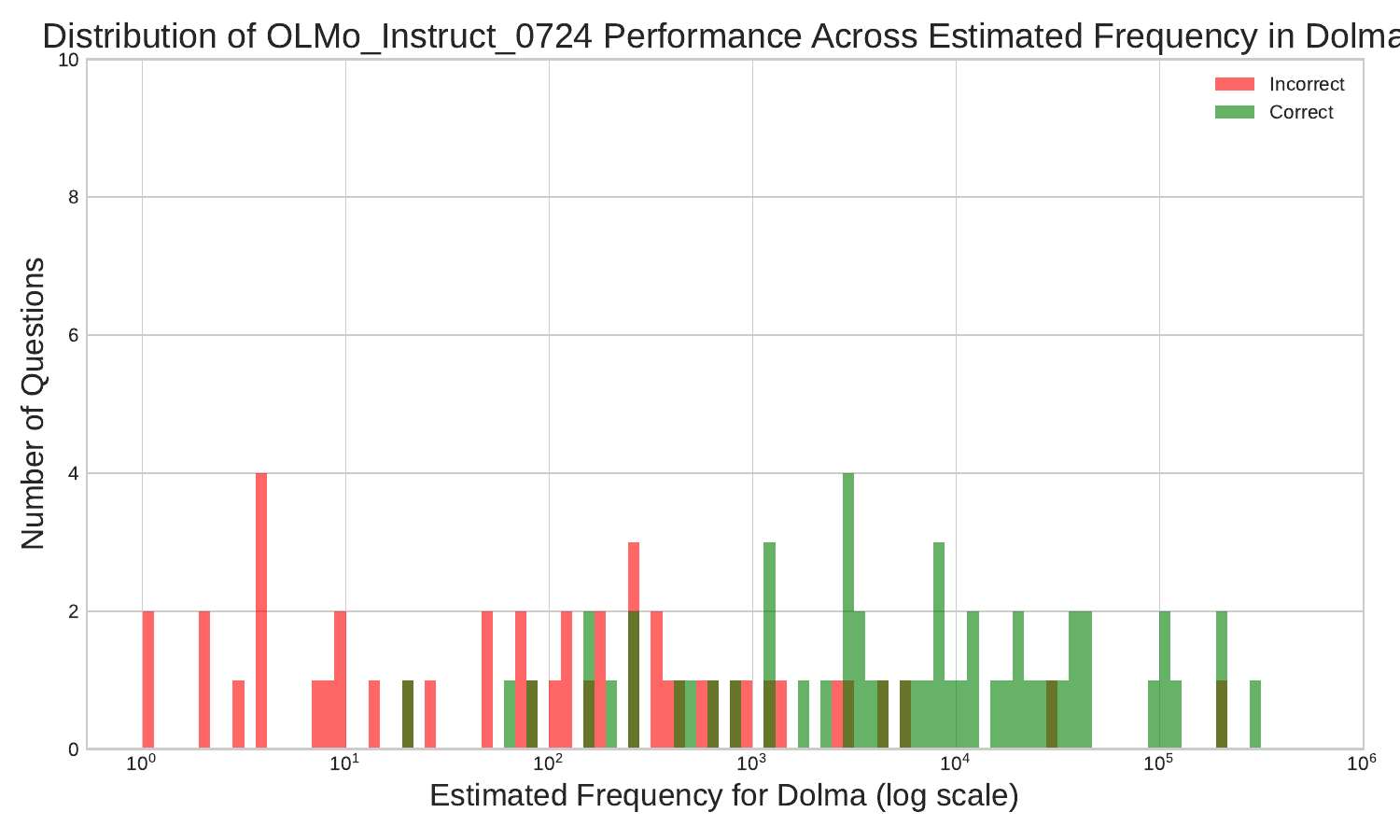}\hfill 
    \caption{Instruction-Tuned Model Correctness vs. Estimated frequency in Pretraining Corpora on \datasetname}
    \label{fig:MIMIC_instructed}
\end{figure*}

In Figure~\ref{fig:correlation_LLaMAs} and ~\ref{fig:correlation_instructed}, each point represents a jargon-expansion pair with the corresponding occurrence in pretraining corpus and accuracy. In Figure~\ref{fig:MIMIC_LLaMA_comparison} and ~\ref{fig:MIMIC_instructed}, each bar shows the number of correct answers alongside the estimated count of occurrence of jargon-expansion documents. 

\paragraph{Frequent Terms Lead to Better Accuracy.}
As shown in all models demonstrated in the Figure \ref{fig:correlation_LLaMAs} \ref{fig:correlation_instructed}, \ref{fig:MIMIC_LLaMA_comparison}, \ref{fig:MIMIC_instructed}, in nearly all models, jargon-expansion pairs with higher frequency in the training corpus show stronger performance. However, one exception is Flan-T5, which performs notably worse on \datasetname, achieving just 37\% overall accuracy. This suggests a limited ability to learn clinical abbreviations, regardless of how frequently they appear.

\paragraph{As model size grows, Rare terms are gradually learned}
Comparing LLaMA variants of different sizes (7B, 13B, 33B, and 65B) in Figures~\ref{fig:correlation_LLaMAs} and \ref{fig:MIMIC_LLaMA_comparison} reveals that larger models maintain decent accuracy even for terms with relatively few examples. Even with the same pretraining corpus and training strategy, the accuracy of models varies a lot on both the CASI and \datasetname. Both Figure ~\ref{fig:correlation_LLaMAs} and ~\ref{fig:MIMIC_LLaMA_comparison} show that as the model size grows, generally more data points have an improved accuracy; mostly frequently appeared pairs have the accuracy increase first and then those less frequently appeared data also demonstrate increased accuracy. The trend is especially eminent across plots in figure \ref{fig:MIMIC_LLaMA_comparison}. Larger models appear more capable of leveraging the limited instances that exist in the corpus, suggesting that additional capacity of the model can improve performance even on rarer items. This also suggests that the performance of the majority of the clinical jargon is not bottlenecked by the existing data in the pretraining corpus, even if they appear a few times, large models can grasp the understanding. 

\begin{figure}[htp]
    \begin{minipage}[t]{0.45\textwidth}
        \centering
        \includegraphics[width=\textwidth]{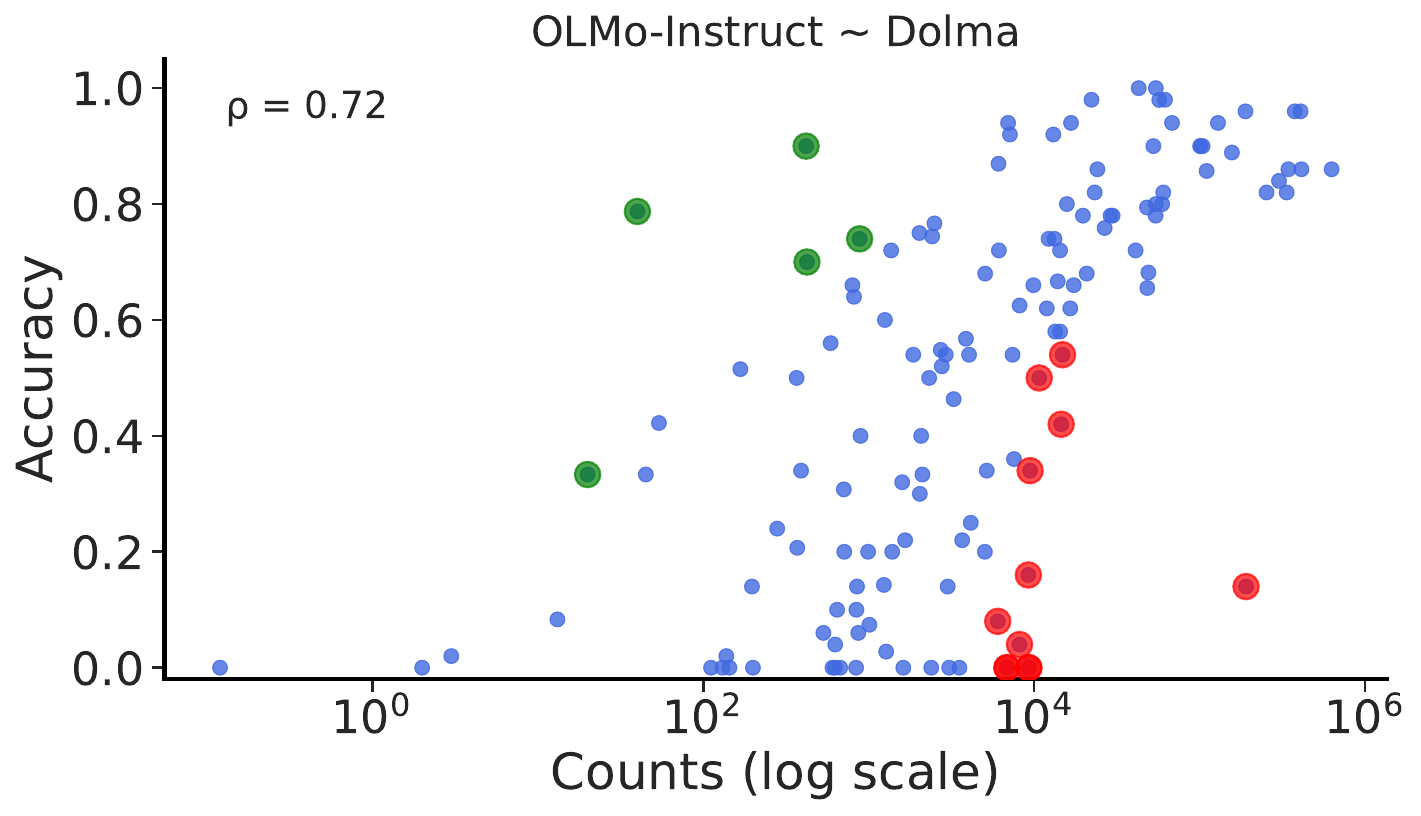}
        \caption{OLMo accuracy vs. Dolma estimated co-occurrence frequency on CASI dataset. Each dot shows a jargon-expansion pair. Green points indicate high-accuracy low-frequency instances (top 5 rows in adjacent table); red points represent low-accuracy despite high-frequency cases (bottom 10 rows).}
        \label{fig:CASI_OLMo_Dolma_outlier}
    \end{minipage}
    \hfill
    \begin{minipage}[t]{0.5\textwidth}
        \centering
        \begin{tabular}{cc}
            \hline
            \textbf{sf} & \textbf{ground\_truth} \\
            \hline
            PAC & post anesthesia care \\
            MOM & multiples of median \\
            MP & metatarsophalangeal \\
            PAC & picture archiving communication \\
            VBG & venous blood gas \\
            \midrule
            OP & operative \\
            AC & acetate \\
            ASA & aminosalicylic acid \\
            CA & carbohydrate antigen \\
            CR & controlled release \\
            SBP & spontaneous bacterial peritonitis \\
            AV & arteriovenous \\
            IR & immediate-release \\
            ALD & adrenoleukodystrophy \\
            LA & long-acting \\
            \hline
        \end{tabular}
        \label{tab:boundary_points}
    \end{minipage}
    
\end{figure}

\begin{table}[t]
\centering
\resizebox{\textwidth}{!}{%
\begin{tabular}{cc|c|cccc}
\toprule
\textbf{abbr} & \textbf{expansion} & \textbf{Estimated Counts}& \multicolumn{4}{c}{\textbf{Accuracy}} \\
& & & 7B & 13B & 33B & 65B \\
\midrule
 PAC & post anesthesia care & 25 & 0.60 & 0.85 & 0.85 & 0.89 \\
 PAC & premature atrial contraction & 200 & 0.72 & 0.84 & 0.96 & 0.98 \\
 CVP & cyclophosphamide, vincristine, prednisone & 381 & 0.76 & 0.72 & 0.86 & 0.92 \\
 MP & metatarsophalangeal & 259 & 0.70 & 0.64 & 0.58 & 0.58 \\
 CVS & cardiovascular system & 1038 & 0.80 & 0.90 & 0.88 & 0.93 \\
 AVR & aortic valve replacement & 4894 & 0.96 & 0.96 & 0.98 & 1.00 \\
 CTA & computed tomographic angiography & 3325 & 0.90 & 0.82 & 0.90 & 0.94 \\
 \midrule
 PA & physician assistant & 64401 & 0.34 & 0.96 & 0.80 & 0.82 \\
 DC & discharge & 29024 & 0.31 & 0.38 & 0.66 & 0.72 \\
 OP & operative & 167626 & 0.28 & 0.78 & 0.28 & 0.40 \\
 PR & progesterone receptor & 17130 & 0.22 & 0.92 & 0.80 & 0.98 \\
 DC & direct current & 120562 & 0.26 & 0.86 & 0.64 & 0.92 \\
 CR & controlled release & 5831 & 0.10 & 0.88 & 0.62 & 0.78 \\
 DT & diphtheria-tetanus & 3885 & 0.02 & 0.18 & 0.36 & 0.18 \\
 SA & saturation & 2123 & 0.00 & 0.00 & 0.25 & 0.25 \\
 AMA & advanced maternal age & 844 & 0.00 & 0.00 & 0.77 & 0.94 \\
 ASA & aminosalicylic acid & 2630 & 0.00 & 0.00 & 0.33 & 0.33 \\
 PD & phosphate dehydrogenase & 1889 & 0.00 & 0.00 & 0.00 & 0.11 \\
\bottomrule
\end{tabular}
}
\caption{Outlier points tracked across different LLaMA model sizes (7B to 65B), with the top rows showing low‑frequency, high‑accuracy cases and the bottom rows showing high‑frequency, low‑accuracy cases in the 7B model.}
\label{tab:llama_outliers}
\end{table}

To complement the correlation results, we identify and inspect outlier abbreviations, those with unexpectedly high accuracy despite low corpus frequency, or conversely, low accuracy despite high frequency. Figure \ref{fig:CASI_OLMo_Dolma_outlier} marks these cases in red (high‑frequency/low‑accuracy) and green (low‑frequency/high‑accuracy), with specific instances listed in the adjacent table of the figure. Meanwhile, we track how the outlier instances changed as the model sizes increased. We look at outlier instances in a LLaMA 7B model and track the accuracy as the model size increases to 65B, as shown in Table \ref{tab:llama_outliers}. For both the high-frequency low-accuracy and the low-frequency high-accuracy instances, the accuracy generally increases as the model size grows.

\begin{figure*}[htp]
    \centering
    \includegraphics[width=.48\textwidth]{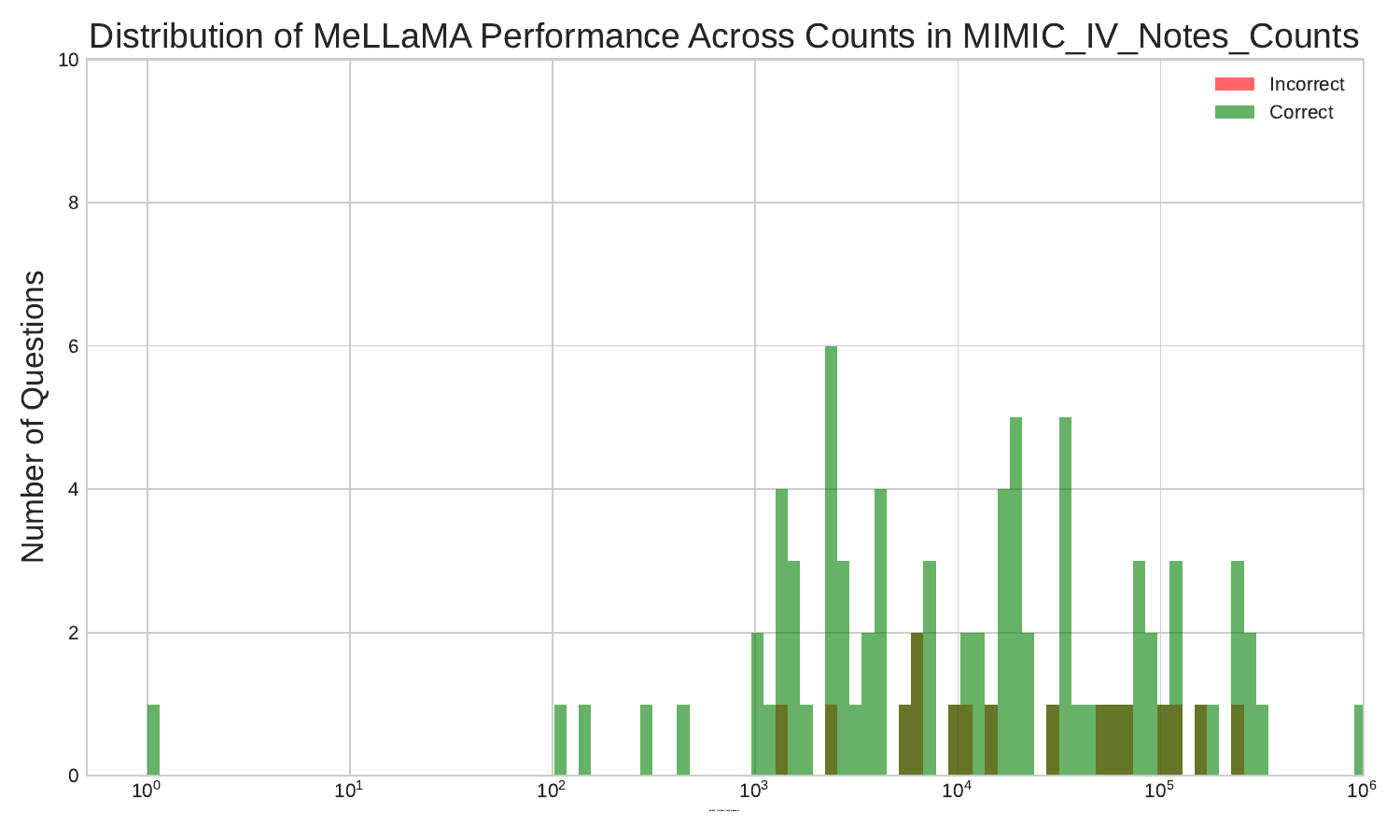}\hfill
    \includegraphics[width=.48\textwidth]{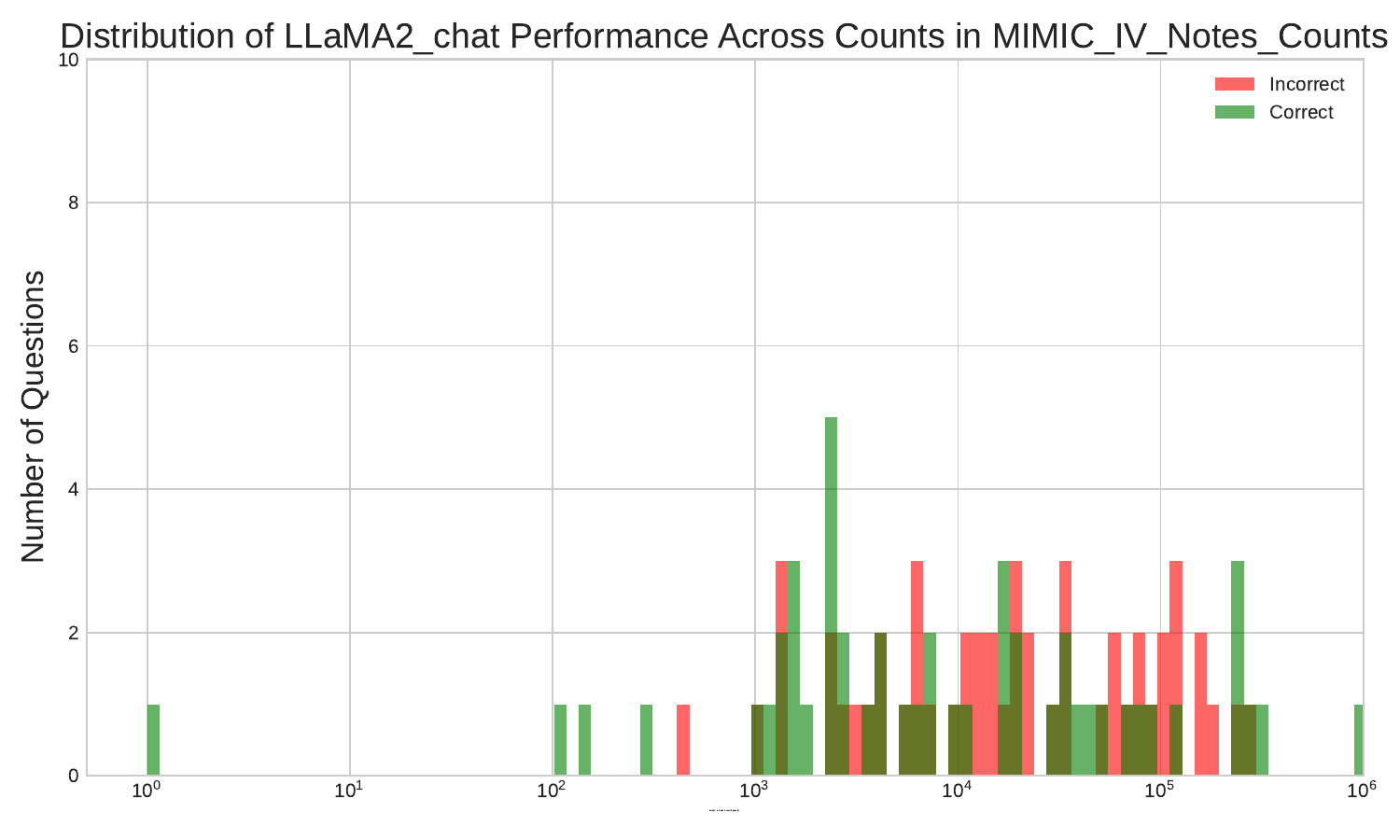}\hfill 
    \includegraphics[width=.48\textwidth]{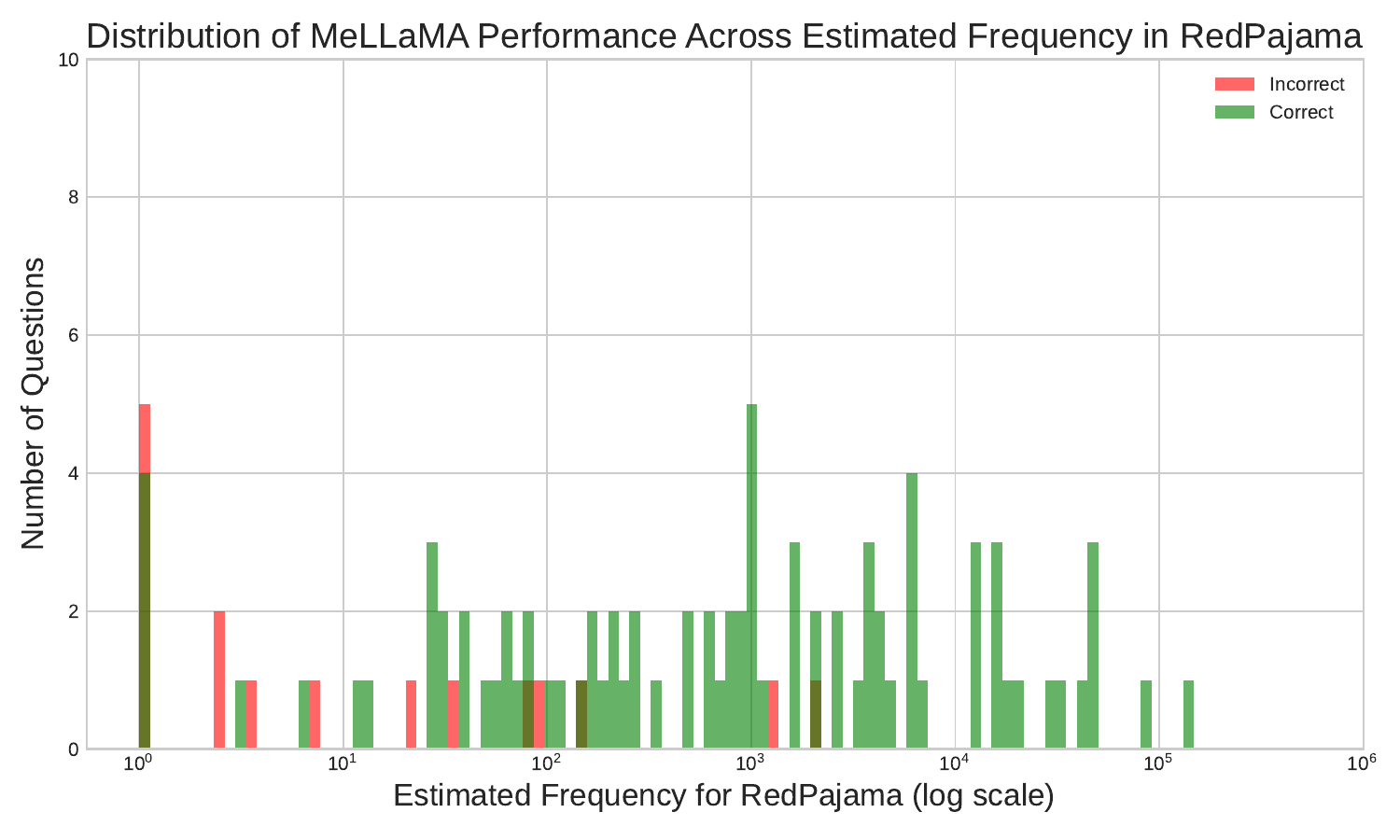}\hfill 
    \includegraphics[width=.48\textwidth]{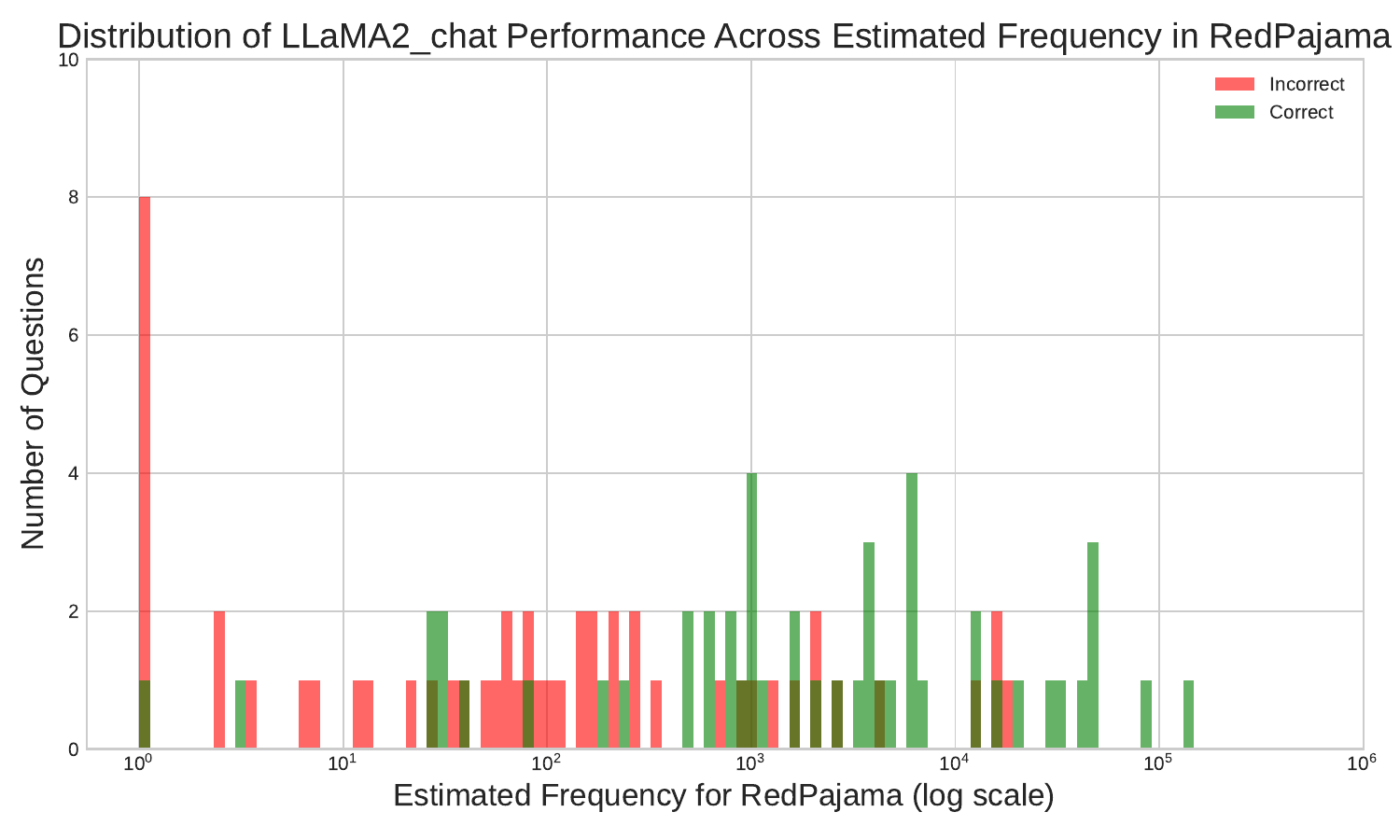}\hfill 
    \caption{Comparsion of MeLLaMA2 and LLaMA2 chat on \datasetname}
    \label{fig:MedLingo_MeLLaMA2_LLaMA2_chat}
\end{figure*}

\section{Disputed Medical Claims Correlation Analysis}
\label{apd:unsupported_medical_claims}

\begin{figure*}[htp]
    \centering
    \includegraphics[width=.48\textwidth]{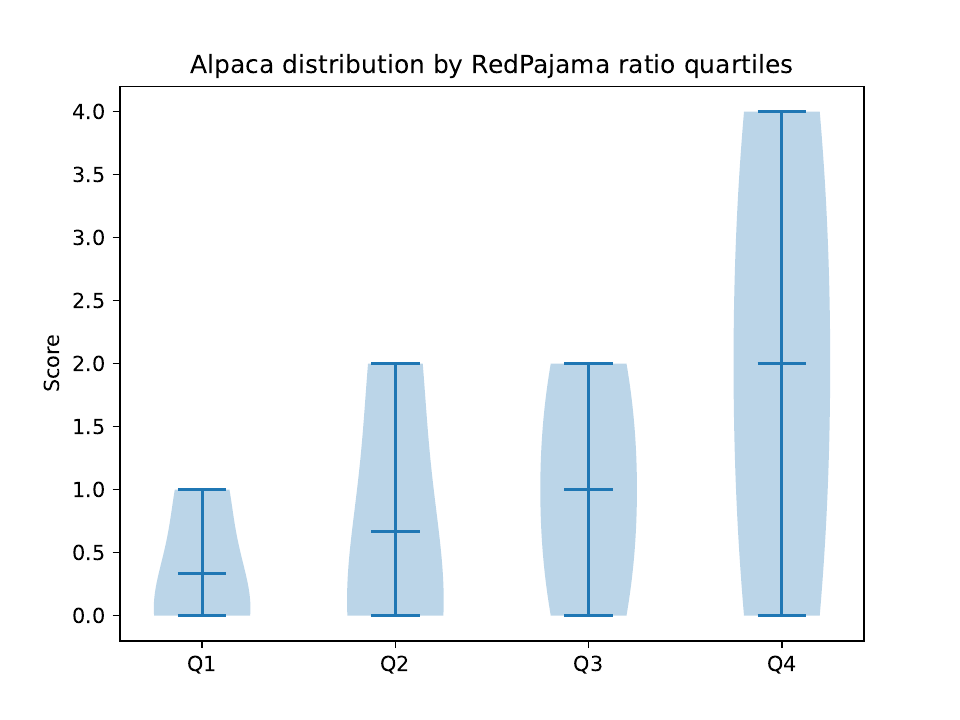}\hfill 
    \includegraphics[width=.48\textwidth]{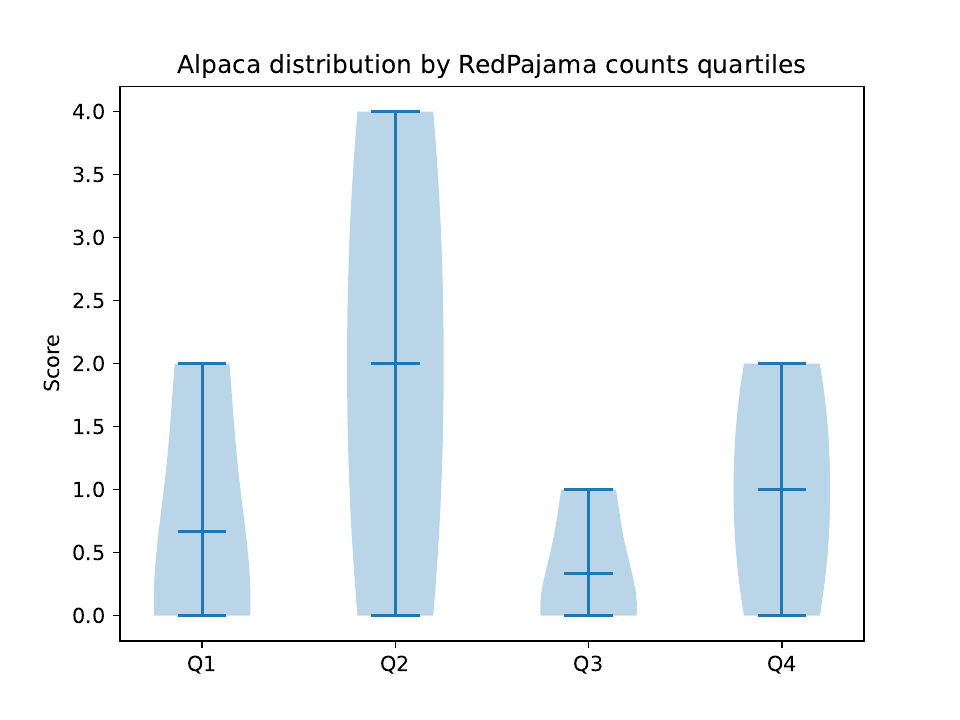}\hfill 
    \caption{Level of Disputed Medical Claims in Alpaca's response across ratio of supportive documents and estimated counts of supportive documents in its pretraining corpora RedPajama}
    \label{fig:unsupported_medical_claims_correlation}
\end{figure*}

We score responses as follows: denial = 0, neutral = 1, and support = 2, and sum these scores for each example across both prompt types for each example. We further compare the correlation between the two metrics, the ratio and the estimated counts of supportive documents, with the score of the disputed medical claims. Although neither metric shows a strong correlation, the ratio exhibits a more meaningful correlation (Spearman correlation $p=0.28$) compared to the estimated counts (Spearman correlation $p=-0.20$). We also observed the same trend across examples. For instance, examples such as "fluroide" with "cancer" and "magnet therapy" with "arthritis" demonstrates high ratios, and both corresponds to a high tendency to output disputed medical claims, while the later pair has a low estimated count for supportive documents. Moreover, stratifying responses into top and bottom 50\% groups further supports that the ratio metric more effectively differentiates levels of disputed medical claims: OLMo's average score increases from 0.33 to 0.67 with ratio-based grouping, while it remains 0.5 with count-based grouping. Similarly, Alpaca's score rises from 0.5 to 1.50 with the ratio split, but with a count split, the trend inverts (1.33 in the bottom half and 0.67 in the top half). Furthermore, stratifying into four quartiles by the metrics also support the findings, as illustrated in Figure \ref{fig:unsupported_medical_claims_correlation}, where ratio-based grouping shows a steady increase in the level of disputed medical claims in the response. These results suggest that the ratio of supportive documents is a more reliable indicator of unsupported medical claims in outputs than the estimated counts, although further work is needed to explore the correlation between the claims in pretraining corpora and the level of disputed medical claims in model outputs at scale.  Evaluation results can be found in the \href{https://github.com/Flora-jia-jfr/diagnosing_our_datasets}{Github Repository}.

\section{Example of Clinical Jargon-Expansion Pairs in \datasetname}

\begin{table}[h]
    \centering
        \resizebox{\textwidth}{!}{%
    \begin{tabular}{ll|ccccll}
 \textbf{Jargon}& \textbf{Expansion}& \textbf{Dolma}&\textbf{C4}&\textbf{RedPajama}&\textbf{MIMIC IV Notes} & \multicolumn{2}{c}{\textbf{Maximum Source Category}}\\
 & & & & & & Source&Percentage\\ 
 \midrule
    AVSS& afebrile, vital signs stable & 12& 0& 0& 10766 & Commercial Health& 7/12\\ 
    BRBPR& bright red blood per rectum & 368& 42& 58& 33533 & Research Publication& 29/99\\ 
    cc& chief complaint & 2740& 269& 1274& 29932 & Other& 25/99\\ 
    CCE& clubbing, cyanosis, and edema & 3& 1& 3& 6698 & Medical Encyclopedia& 2/3\\ 
    chole& cholecystostomy & 18& 1& 6& 4193 & Medical Encyclopedia& 6/18\\ 
    HKS& heel-knee-shin test & 1& 0& 0& 6457 & Medical Encyclopedia& 1/1\\ 
    HLD& hyperlipidemia & 1393& 191& 711& 74585 & Research Publication& 54/100\\ 
    HSM& hepatosplenomegaly & 271& 35& 84& 55181 & Research Publication& 49/100\\ 
    MMM& moist mucous membrane & 1& 0& 0& 168801 & Clinician Forum& 1/1\\ 
    NBS& normal bowel sounds & 72& 2& 21& 1372 & Other& 44/72\\
 NC&normocephalic & 50& 10& 33& 114379 & Other& 16/50\\
 NGTD&no growth to date & 26& 35& 11& 13043 & Research Publication& 4/7\\
 NVI&neurovascularly intact & 7& 0& 0& 5546 & Medical Encyclopedia& 3/7\\
 QPM&every afternoon & 8& 1& 2& 157805 & Medical Encyclopedia& 3/7\\
 RPRNR&nonreactive RPR (Rapid Plasma Reagin) & 0& 0& 0& 1317 & Clinician Forum& 0/0\\
 sp&status post & 48& 396& 323& 73993 & Other& 9/25\\
 Tc&Tympanic Membrane Temperature & 3& 0& 2& 14508 & Research Publication& 2/3\\
 Utox&urine toxicology screen & 256& 35& 101& 3162 & Research Publication& 11/13\\ 
    \end{tabular}
    }
    \caption{Jargon-Expansion pairs in \datasetname{} that all models on open-source pretraining corpora fail}
    \label{tab:jargons_fail}
\end{table}

\begin{table}[h]
    \centering
    \resizebox{\textwidth}{!}{%
    \begin{tabular}{ll|ccccll}
 \textbf{Jargon}& \textbf{Expansion}& \textbf{Dolma}&\textbf{C4}&\textbf{RedPajama}&\textbf{MIMIC IV Notes} & \multicolumn{2}{c}{\textbf{Maximum Source Category}}\\
 & & & & & & Source&Percentage\\ 
 \midrule
    Abx& antibiotics& 36701& 7381& 13364& 28046 & Patient Forum&60/99\\ 
    amio& Amiodarone& 2945& 154& 951& 2452 & Research Publication&29/99\\ 
    brady& bradycardia& 2972& 482& 1641& 5175 & Research Publication&30/97\\ 
    bx& biopsy& 5341& 708& 2493& 11592 & Research Publication&33/100\\ 
    coag& coagulation& 9441& 2796& 6477& 19577 & Research Publication&29/94\\ 
    ddx& differential diagnosis& 11404& 10860& 3727& 6774 & Medical Encyclopedia&24/100\\ 
    DM2& Type 2 diabetes& 23182& 1087& 7211& 34022 & Research Publication&47/100\\ 
    etoh& alcohol& 120236& 8570& 31384& 70849 & Research Publication&46/97\\ 
    FHx& family history& 1710& 343& 629& 2209 & Research Publication&42/99\\ 
    fx& fracture& 6283& 3220& 2191& 16449 & Other&24/100\\
 GBM&glioblastoma& 204479& 29267& 82529& 1824 & Research Publication&66/99\\
 hd&hemodialysis& 314956& 4101& 148309& 124976 & Research Publication&80/100\\
 HTN&hypertension& 100600& 10102& 28103& 285064 & Research Publication&53/99\\
 LCx&left circumflex artery& 19521& 1327& 4146& 23790 & Research Publication&87/99\\
 MTX&methotrexate& 208569& 9825& 48585& 17704 & Research Publication&61/100\\
 nl&normal limits& 269& 16& 179& 92981 & Research Publication&43/68\\
 NS&normal saline& 15803& 980& 5902& 49587 & Research Publication&70/99\\
 osm&osmolarity& 1165& 290& 486& 2549 & Research Publication&53/98\\
 RUQUS& right upper quadrant ultrasound& 18& 1& 13&6149 & Clinician Forum&2/4\\
 subq& subcutaneous& 26097& 3432& 6082&2912 & Commercial Health&23/100\\
 Sx& symptoms& 42050& 2080& 16177&20118 & Patient Forum&27/89\\
 trach& tracheotomy& 44240& 15624& 16864&18097 & Personal Blog&36/100\\
 Vanc& vancomycin& 2946& 311& 1188&33783 & Research Publication&56/100\\
 vfib& ventricular fibrillation& 3948& 472& 759&1439 & Other&19/100\\ 
    \end{tabular}
    }
    \caption{Jargon-Expansion pairs in \datasetname{} that all models on open-source pretraining corpora succeed}
    \label{tab:jargons_succeed}
\end{table}

Tables \ref{tab:jargons_fail} and \ref{tab:jargons_succeed} list the jargon–expansion pairs from the \datasetname{} dataset on which models with open-source pretraining corpora (including LLaMA models of varying size, Alpaca, Flan T5, and OLMo Instruct) fail or succeed, respectively. Although the counts of these terms in MIMIC IV notes are similar, models tend to fail on pairs that are rarely represented in the pretraining corpora and succeed on those that are well represented. For each pair, we also report the majority source category and its percentage based on the total documents examined. Among the 24 pairs that all models succeed on, 16 are predominantly sourced from peer-reviewed research publications. In contrast, for the 18 pairs that all models fail on, only 7 are primarily from research publications; 5 are mainly from medical encyclopedias/dictionaries, 4 from other sources, 1 from clinical forums, and 1 from commercial health sources. This suggests that clinical jargon supported by fewer documents tends to originate from informal sources rather than academic literature, potentially offering less clinical contextual information for effective model learning.

\end{document}